\documentclass{article}

\usepackage{arxiv}

\usepackage{cite}
\usepackage{amsmath,amssymb,amsfonts}
\usepackage{algorithmic}
\usepackage{graphicx}
\usepackage{textcomp}
\usepackage{amsmath}
\usepackage{mathtools}
\usepackage{amsmath}
\usepackage{xcolor}

\usepackage[utf8]{inputenc} 
\usepackage[T1]{fontenc}    
\usepackage{hyperref}       
\usepackage{url}            
\usepackage{booktabs}       
\usepackage{amsfonts}       
\usepackage{nicefrac}       
\usepackage{microtype}      
\usepackage{lipsum}
\usepackage{fancyhdr}

\newcommand{\lvec}{\overleftarrow}    
\newcommand{\rvec}{\overrightarrow}    

\title{DeepHealth: Review and challenges of artificial intelligence in health informatics}

\author{
  Gloria Hyunjung Kwak\\
  Department of Computer Science and Engineering\\
  The Hong Kong University of Science and Technology\\
  Hong Kong \\
  \texttt{hkwak@cse.ust.hk} \\
   \And
  Pan Hui \\
  Department of Computer Science and Engineering\\
  The Hong Kong University of Science and Technology\\
  Hong Kong \\
  \texttt{panhui@cse.ust.hk} 
}

\begin{document}

\maketitle

\begin{abstract}
Artificial intelligence has provided us with an exploration of a whole new research era. As more data and better computational power become available, the approach is being implemented in various fields. The demand for it in health informatics is also increasing, and we can expect to see the potential benefits of its applications in healthcare. It can help clinicians diagnose disease, identify drug effects for each patient, understand the relationship between genotypes and phenotypes, explore new phenotypes or treatment recommendations, and predict infectious disease outbreaks with high accuracy. In contrast to traditional models, recent artificial intelligence approaches do not require domain-specific data pre-processing, and it is expected that it will ultimately change life in the future. Despite its notable advantages, there are some key challenges on data (high dimensionality, heterogeneity, time dependency, sparsity, irregularity, lack of label, bias) and model (reliability, interpretability, feasibility, security, scalability) for practical use.
\par 
This article presents a comprehensive review of research applying artificial intelligence in health informatics, focusing on the last seven years in the fields of medical imaging, electronic health records, genomics, sensing, and online communication health, as well as challenges and promising directions for future research. We highlight ongoing popular approaches' research and identify several challenges in building models.
\end{abstract}


\section{Introduction}
\label{sec:introduction}

\par Artificial intelligence (AI) has become trends in recent years, opening up a whole new research era in various fields. Demand for AI in health care has increased in academia and industry, and the potential benefits of its applications have been demonstrated. Previous studies have attempted to implement AI approaches on medical images, electronic health records (EHRs), molecular characteristics, and a variety of lifestyles \cite{miotto2017deep,ravi2016deep, litjens2017survey,lyman2016biomarker,collins2015new,shickel2017deep,angermueller2016deep,esteva2019guide,yin2019systematic,arXiv190709475L,kumar2019deep,havaei2016deep,luo2016big,Diao2018Biomedical,gawehn2016deep,eraslan2019deep,pastur2016deep,Yablowitz2018Areview,Patel2012review,cosgriff2019critical}. Researchers used data aggregated from multiple data sources to train models that mimic what clinicians do when they see patients and helped decision support through results and interpretations. It included how to read clinical images, predict outcomes, find the relationship between genotype and phenotype or phenotype and disease, analyze treatment responses, and track lesions or structural changes (ex. hippocampal volume reduction). Moreover, the prediction study (ex. disease or readmission) and correlation and pattern identification study have been extended to an early warning system with risk scores and global pattern studies and population health care (ex. providing predictive care for the entire population).

\par Among various observation methods, ranging from comparatively simple statistical projections to machine learning (ML) and deep learning (DL) algorithms, several architectures stood out in popularity. Researchers started with a mass univariate analysis model such as t-test, F-test, and chi-square test to prove the contrast of a null hypothesis. They continued to the ML methods with feature extraction for classification, prediction, de-identification, and treatment recommendation. Moreover, DL algorithms gave researchers many choices of experiments with neuro-inspired techniques to generate optimal weights, abstract high-level features on its own, and extract information factors non-manually, resulting in more objective and unbiased classification results \cite{schmidhuber2015deep, srivastava2014dropout, lecun2015deep}.

\par Deep learning in health informatics has shown many advantages: it can be trained without a priori and feature extraction from human experts, which combats the lack of labeled data and burden on clinicians, and has expanded the possibilities of research in health informatics. First of all, in medical imaging, researchers dealt with data complexity, overlapped detection target points, and 3- or 4-dimensional medical images, and provided sophisticated and elaborative outcomes with data augmentation, un-/semi-supervised learning, transfer learning, and multi-modality architectures \cite{nie20163d, xu2016multimodal,chang2017unsupervised,samala2016mass,yan2016multi}. Second of all, the nonlinear relationship between variables was discovered. It helped clinicians, and patients with an objective and personalized definition of disease and solutions since decisions are made up of data rather than human intervention and models divide the cohort into subgroups according to their clinical information. As an example, DNA or RNA sequences were studied to identify gene alleles and environmental factors that contribute to diseases, investigate protein interactions, understand higher-level processes (phenotype), find similarities between two phenotypes, design targeted personalized therapies, and more \cite{leung2015machine}. In particular, deep learning algorithms were implemented to predict the splicing activity of exons, the specificities of DNA-/RNA- binding proteins, and DNA methylation \cite{xiong2015human,alipanahi2015predicting,angermueller2016accurate}. Third, it demonstrated its usefulness with discovering new phenotypes and subtypes, delivering personalized patient-level and real-time level risk prediction rather than regularly scheduled health screenings and guiding treatment \cite{tomavsev2019clinically,Davis2017,Seymour2019,Knaus2019,Prendecki9,Hoste2016,Goldstein2017,Bamgbola2016,wang2018supervised}. It shows its potential, especially when predicting rapid developing diseases such as acute renal failure \cite{akigoogle}. Fourth, it was used for first-time inpatients, transferred patients, weak healthcare infrastructure patients, and outpatients without chart information \cite{discharge2002,barth2011biometric}. For example, portable neurophysiological signals such as Electroencephalogram (EEG), Local Field Potentials (LFP), Photoplethysmography (PPG) \cite{jindal2016adaptive, nurse2016decoding}, accelerometer data from above ankle and mobile apps were used to monitor individual health status, to predict freezing from Parkinson's disease, rheumatoid arthritis, chronic diseases such as obesity, diabetes and cardiovascular disease to provide health information before hospital admission and to prepare the emergency intervention. Besides, mobile health technologies for resource-poor and marginalized communities were also studied with reading X-ray images taken by a mobile phone \cite{cao2016improving}. Clinical notes studies were aimed to study how summarization notes express reliable, effective, and accurate information timely, comparing the information with medical records. Finally, disease outbreaks, social behavior, drug/treatment review analysis, and research on remote surveillance systems have also been studied to prevent disease, prolong life, and monitor epidemics \cite{phan2015social,garimella2016social,bodnar2014ground,tuarob2014ensemble,de2016hayfever,Alimova2017inps,chae2018predicting}. 
\par
\begin{figure}[!htbp]
    \centering
    \centerline{
    \includegraphics[scale=0.4]{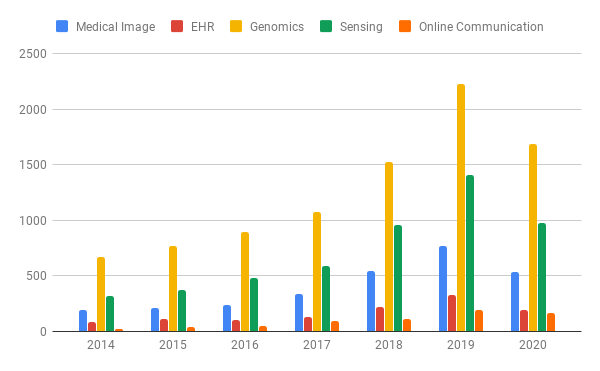}
    \includegraphics[scale=0.4]{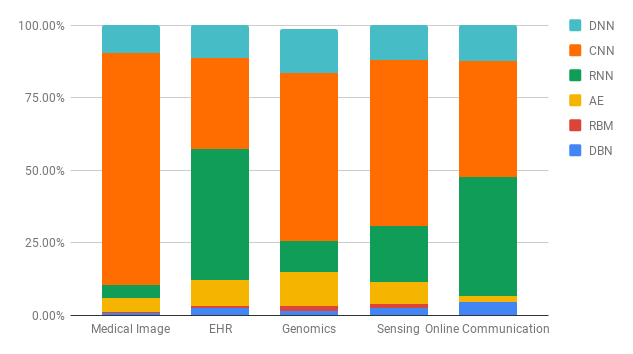}} 
    \caption{Left: Distribution of published papers that use artificial intelligence in subareas of health informatics from PubMed, Right: Percentage of most used deep learning methods in health informatics. (DNN: Deep Neural Network, CNN: Convolutional Neural Network, RNN: Recurrent Neural Network, AE: Autoencoder, RBM: Restricted Boltzmann Machine, DBN: Deep Belief Network)}
    \label{Img:Fig0}
\end{figure}

 In the sense of trust and expectation, the number of papers snowballed, illustrated in Fig.~\ref{Img:Fig0}. The number of hospitals that have adopted at least a basic EHR system drastically increased. Indeed, according to the latest report from the Office of the National Coordinator for Health Information Technology (ONC), nearly over 75\% of office-based clinicians and 96\% of hospitals in the United States using an EHR system, nearly all practices have an immediate, practical interest in improving the efficiency and use of their EHRs \cite{birkhead2015uses, henry2016adoption}. With the rapid development of imaging technologies (magnetic resonance imaging, positron emission tomography, computed tomography), wearable sensors, genomic technologies (microarray, next-generation sequencing), information about patients can now be more readily acquired. Thus far, deep learning architectures have developed with computation power support in Graphics Processing Units (GPUs), which have significantly impacted practical uptake and acceleration of deep learning. Therefore, plenty of experimental works have implemented its models for health informatics, reaching clinical alternative technology levels in some areas. 

\par 
Nevertheless, the application of deep learning to health informatics raises several critical challenges that need to be resolved, including data informativeness (high dimensionality, heterogeneity, multi-modality), lack of data (missing values, class imbalance, expensive labeling, fairness, and bias), data credibility and integrity, model interpretability and reliability (tracking and convergence issues as well as overfitting), feasibility, security and scalability. 
    
\par In the following sections of this review, we examine a rapid surge of interest in recent health informatics studies including bioinformatics, medical imaging, electronic health records, sensing, and online communication health, with practical implementations, opportunities, and challenges.

\section{Models Overview}
\label{sec:modeloverview}
\par This section reviews the most common models used in the studies reviewed in this paper. There are various artificial intelligence designs available today, so only a brief introduction to the main base models applied to health informatics is presented. We begin by introducing some common non-deep learning models used in many studies to compare or combine with deep learning models. Deep learning architectures are subsequently reviewed, including convolutional neural network (CNN), recurrent neural network (RNN), graph neural network (GNN), autoencoder (AE), restricted Boltzmann machine (RBM), deep belief network (DBN), and their variants with transfer learning, attention learning, and reinforcement learning.

    \subsection{Support Vector Machine}
    Support vector machine (SVM) aims to define an optimal hyperplane which can distinguish groups from each other. In a training phase, when data are linearly separable \cite{vapnik1995svm}, SVM finds a hyperplane with the longest distance between support vectors of each group (ex. disease case and healthy control groups). If training data are not linearly separable, SVM can be extended to a soft-margin SVM and kernel-trick methods \cite{chen2004support, shawe2000support}. 
    \par For an original SVM, with training data $n$ points of the form $\{\mathbf x_i, y_i\}\ i = 1,\ldots,l$, $\mathbf x_i \in \mathbb{R}^n$, and $y_i \in \{ -1, 1\}$ where $y_i$ are either 1 or -1 and $x_i$ is a $n$-dimensional real vector, minimizing ${\|\mathbf w\|}$ is aimed subject to 
    \begin{equation}
      y_i(\mathbf x_i \mathbf w + b) - 1 \geq 0, \forall i
    \end{equation}
    for all $\mathbf i = 1,\ldots,l$. Unlike a hard-margin algorithm, an extended SVM with a soft-margin introduces a different minimizing problem (hinge loss function) with a trade-off parameter (Equation 2). Having a regularization term ${\|\mathbf w\|^2}$ and a small value parameter makes data can be finally linearly classifiable.
    \begin{equation}
      [\frac{1}{n} \displaystyle\sum_{i=1}^l max(0, 1- y_i(\mathbf x_i \mathbf w - b))] + \lambda {\|\mathbf w\|^2}
    \end{equation}
    \par For another option for non-linearly classifiable data to become linearly separable, kernel methods helps with a feature map $\varphi : \mathcal {X} \rightarrow \mathcal {V}$ which satisfies $k(x, x') = \langle \varphi(x),\varphi(x') \rangle_\nu$ and two of the popularly used methods are polynomial kernel $k(x_i,x_j)=(x_ix_j)^2$
    and Gaussian radial basis kernel $k(x_i,x_j)=\exp(-\gamma\|x_i-x_j\|^2)$. Kernel-trick methods seek a certain dimension which helps data can be linearly separable.
    
    \subsection{Matrix/Tensor Decomposition}
\begin{figure}[!htbp]
    \centering
    \centerline{
    \includegraphics[scale=0.14]{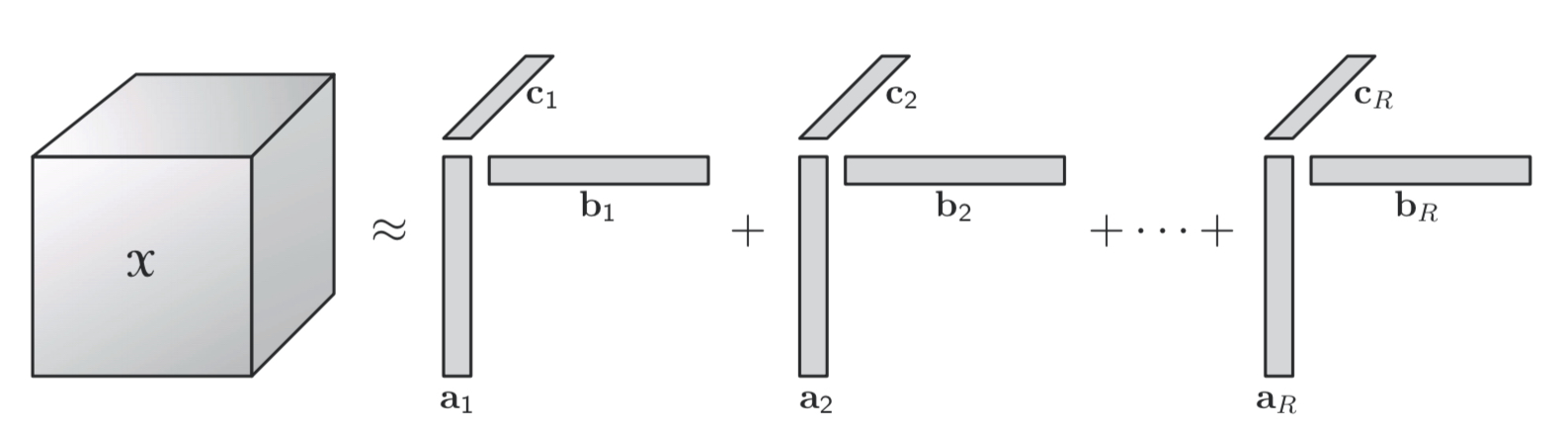}}
    \caption{CP decomposition of a three-way array \cite{kolda2006multilinear}}
    \label{Img:Fig6}
    \end{figure}

    A tensor is a multi-dimensional array. More formally, an $n$-way or $n^t{}^h$-order tensor is an element of a tensor product of $n$ vector spaces, each of which has its coordinate system. A first-order tensor is a vector, and a second-order tensor is a matrix. So, normally second-order tensor decomposition is called matrix decomposition, and three or higher-order tensor decomposition is called tensor decomposition. One of the tensor decompositions is CANDECOMP/PARAFAC (CP) decomposition, and a third-order tensor is factorized into a sum of component rank-one tensors, as shown in Fig.~\ref{Img:Fig6}. For a third-order tensor $\mathbf{X} \in \mathbb{R}^I{^\times{}^J{^\times{}^K}}$, it can be also stated in Equation 3 and 4 for a positive integer $\mathbb{R}$ and $a_r \in \mathbb{R}^I$, $b_r \in \mathbb{R}^J$, $c_r \in \mathbb{R}^K$ \cite{kolda2006multilinear}. 

    \begin{equation}
      \mathbf X \approx [\mathbb{A}, \mathbb{B}, \mathbb{C} ] \equiv
      \displaystyle\sum_{r=1}^R a_r \circ b_r \circ c_r
    \end{equation}
    
    \begin{equation}
    \begin{gathered}
      {\mathbf x_i{_j{}_k} \approx 
      \displaystyle\sum_{r=1}^R a_i{}_r b_j{}_r c_k{}_r 
      \;\, } \\ {for \; \mathbf i = 1,\ldots,I,\,  \mathbf j = 1,\ldots,J,\,  \mathbf k = 1,\ldots,K } 
    \end{gathered}
    \end{equation}

    \subsection{Word Embedding}
\begin{figure}[!htbp]
    \centering
    \centerline{
    \includegraphics[scale=0.23]{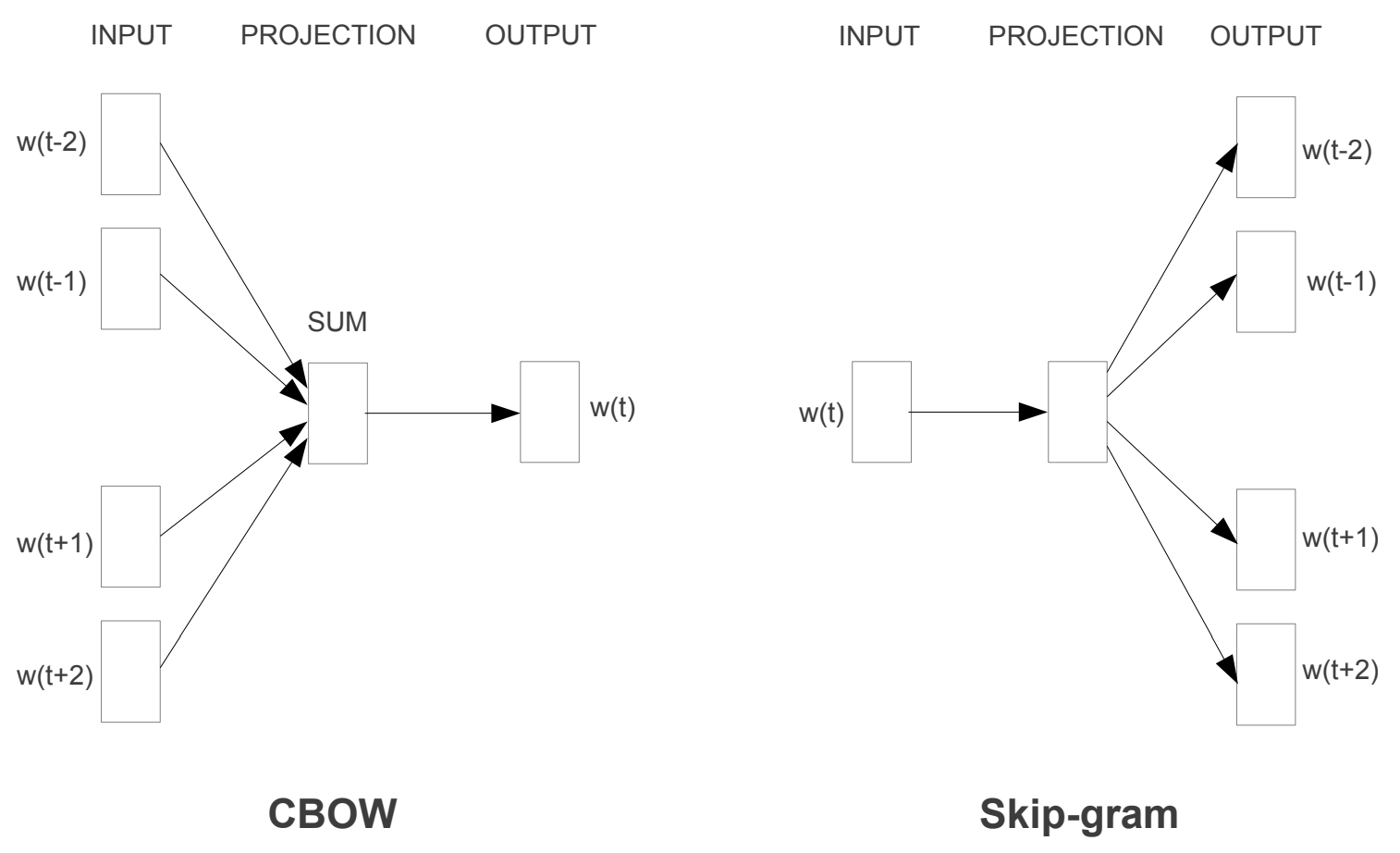}}
    \caption{CBOW and Skip-gram \cite{mikolov2013efficient}}
    \label{Img:Fig5}
    \end{figure}

    Word embedding is a technique to map words to vectors with real numbers, and word2vec is a group of models to produce word embedding \cite{wiki:word2vec}. It allows a model to have more informative and condensed features. Conceptually, with similarity and co-occurrence, words are mapped to a binary space with many dimensions first and then to a continuous vector space with much lower dimensions. Word2vec can be introduced with two distributed representations of words: continuous bag-of-words (CBOW) and skip-gram. CBOW predicts a current word with surrounding context words, and skip-gram uses a present word to predict a surrounding window of context words as Fig.~\ref{Img:Fig5} \cite{mikolov2013efficient}.

    \subsection{Multilayer Perceptron}
    Perceptron is an ML algorithm that researchers refer to as the first online learning algorithm. Multilayer perceptron (MLP) is a feedforward neural network that has perceptrons (neurons) for each layer \cite{lecun2015deep}. When a model has three layers, the minimum amount of layers, the network is called either vanilla or shallow neural network, and when it is deeper than three layers, it is called a deep neural network (DNN). In the case of $n$ layers, the first layer is an input layer (when 1-dimensional data are trained, a list of voxels' intensity corresponds to input data), the last layer is an output layer, and $n-2$ layers are hidden layers. In contrast to SVM, MLP does not require prior feature selection, since it combines some features and finds optimal ones by itself. 
    \par As an online learning-based algorithm that trains data line by line (sample by sample), the model compares the expected value and the labeled value for every sample. The difference between the predicted value and the given labeled value reflects the cost or error. The amount and direction of weights are changed with backpropagation, toward minimizing the error and preventing overfitting with dropout \cite{srivastava2014dropout, lecun2015deep, rumelhart1988learning}.

    \subsection{Convolutional Neural Networks}
\begin{figure}[!htbp]
    \centering
    \centerline{
    \includegraphics[scale=0.3]{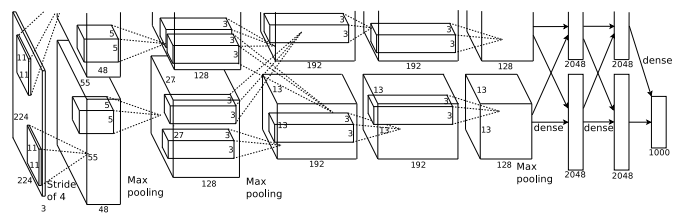}
    \includegraphics[scale=0.2]{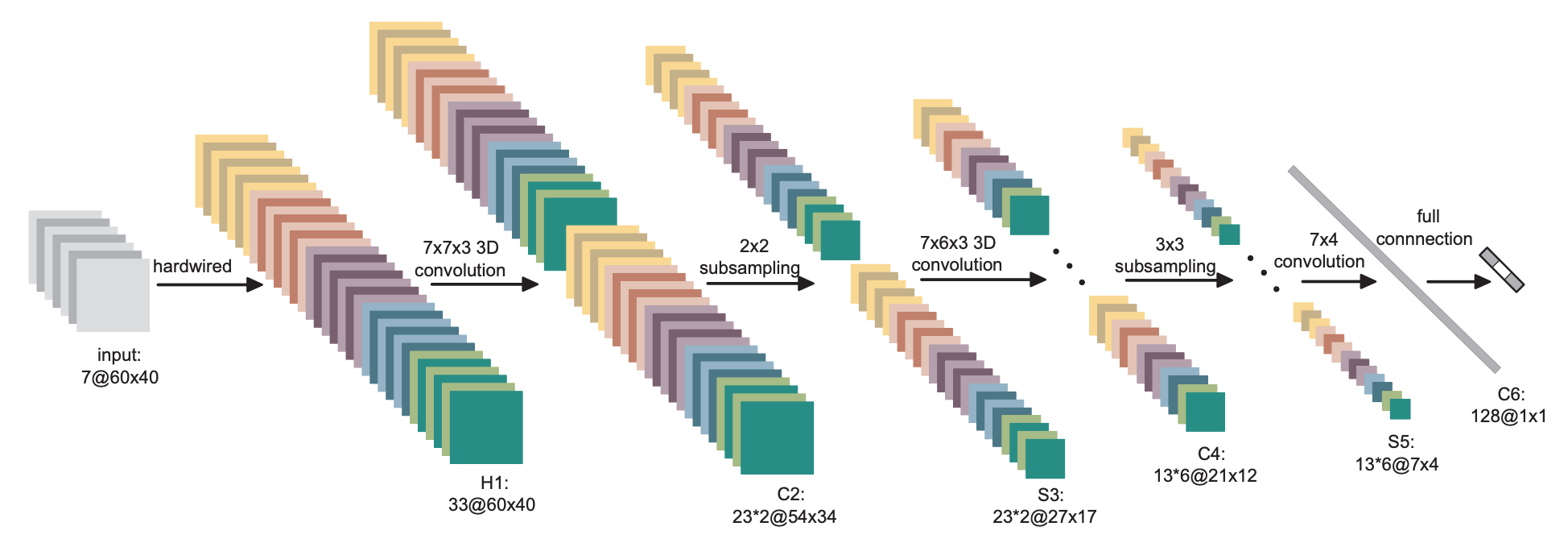}}
    \caption{Left: The architecture of AlexNet (2D CNN) \cite{krizhevsky2012imagenet}, Right: The architecture of 3D CNN \cite{ji20123d}}
    \label{Img:Fig3}
    \end{figure}

    Convolutional neural network (CNN) is an algorithm inspired by biological processing of the animal visual cortex \cite{lecun2015deep, lecun1998gradient, krizhevsky2012imagenet}. Unlike the original fully connected neural network, the algorithm eventually implements how the animal visual cortex works, with convolutional layers with shared sets of 2-dimensional weights for 2-dimensional CNN (2D CNN) cases. And it recognizes the spatial information and pooling layers to filter comparatively more important knowledge and only transmit concentrated features (Fig.~\ref{Img:Fig3}) \cite{lecun1998gradient, hubel1962receptive}. As other deep learning algorithms have a way of preventing overfitting, CNN classifies whether images have specific labels that they look for or not with convolutional and pooling layers. 3D CNN uses three-dimensional weights (Fig.~\ref{Img:Fig3}) \cite{ji20123d}, and 2.5D CNN uses two-dimensional weights with multi-angle learning architectures.

    \subsection{Recurrent Neural Networks}
    
\begin{figure}[!htbp]
    \centering
    \centerline{
    \includegraphics[scale=0.3]{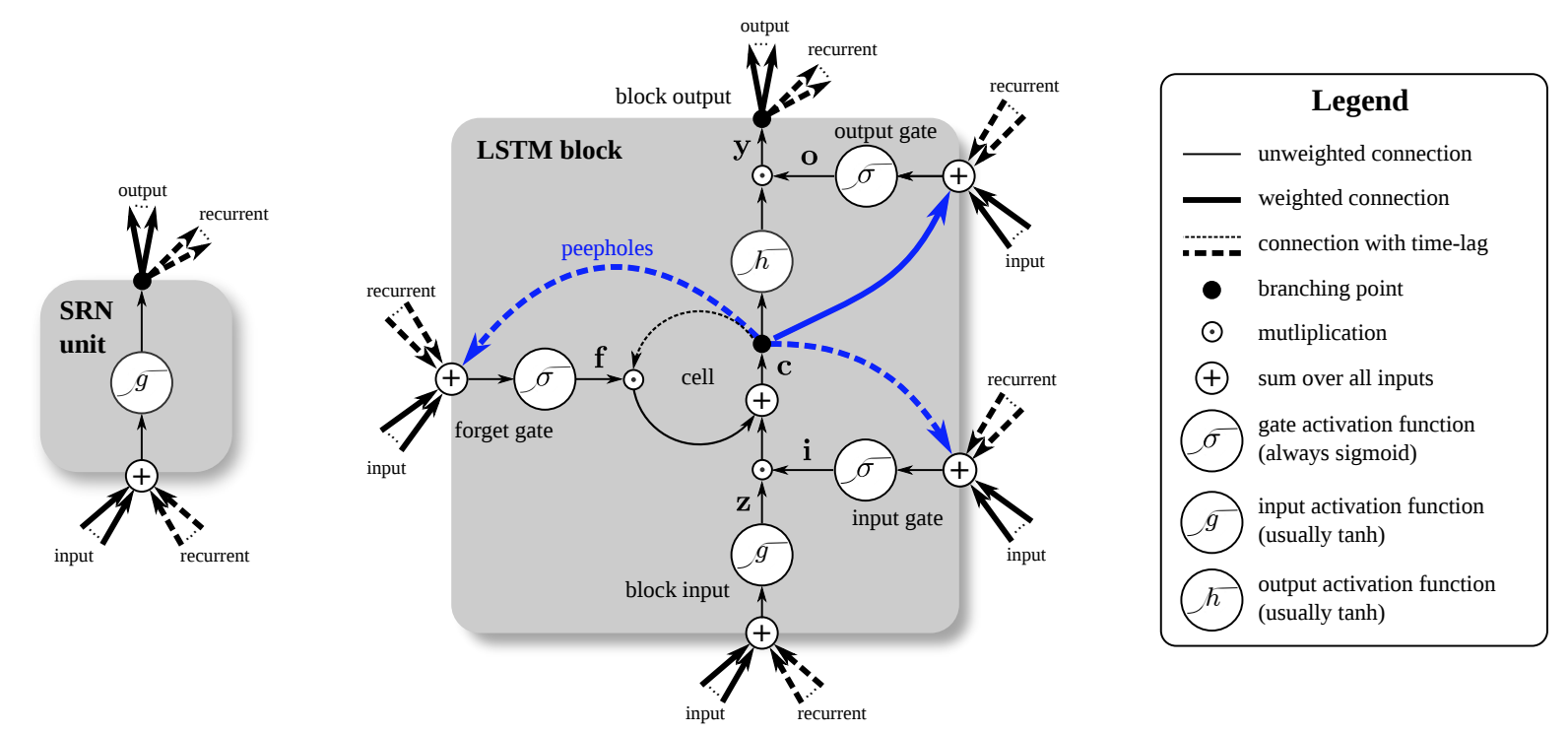}}
    \caption{Left: Detailed schematic of the Simple Recurrent Network (SRN) unit, Right: The architecture of a Long Short-Term Memory block as used in the hidden layers of a recurrent neural network \cite{greff2016lstm}}
    \label{Img:Fig4}
    \end{figure}

    Recurrent neural network (RNN) is a class of ANN specialized for streams of data such as time-series data and natural language \cite{williams1989learning, cho2014learning, greff2016lstm, DBLP:journals/corr/abs-1103-0398}. RNN operates by sequentially updating a hidden state based on the activation of the current input $x$ at the time and the previous hidden state $h_t{}_-{}_1$. Likewise, $h_t{}_-{}_1$ is updated from $x_t{}_-{}_1$ and $h_t{}_-{}_2$, and each output values are dependent on the previous computations. Even though RNN showed significant performance on temporal data, RNN had limitations in vanishing gradient and exploding gradient \cite{bengio1994learning}. For that, RNN variants have been developed, and some well-known examples are long short-term memory (LSTM) and gated recurrent units (GRU) networks, which addressed these problems by capturing long-term dependencies (Fig.~\ref{Img:Fig4}) \cite{hochreiter1997long}. RNN is used not only for time-series predictive modeling but also for missing data imputation.

    \subsection{Autoencoders}
\begin{figure}[!htbp]
    \centering
    \centerline{
    \includegraphics[scale=0.2]{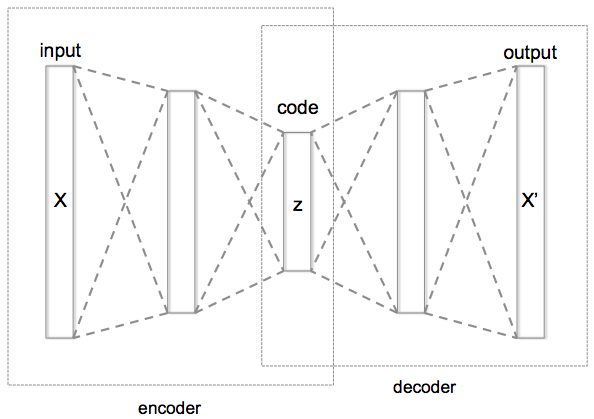}
    \includegraphics[scale=0.25]{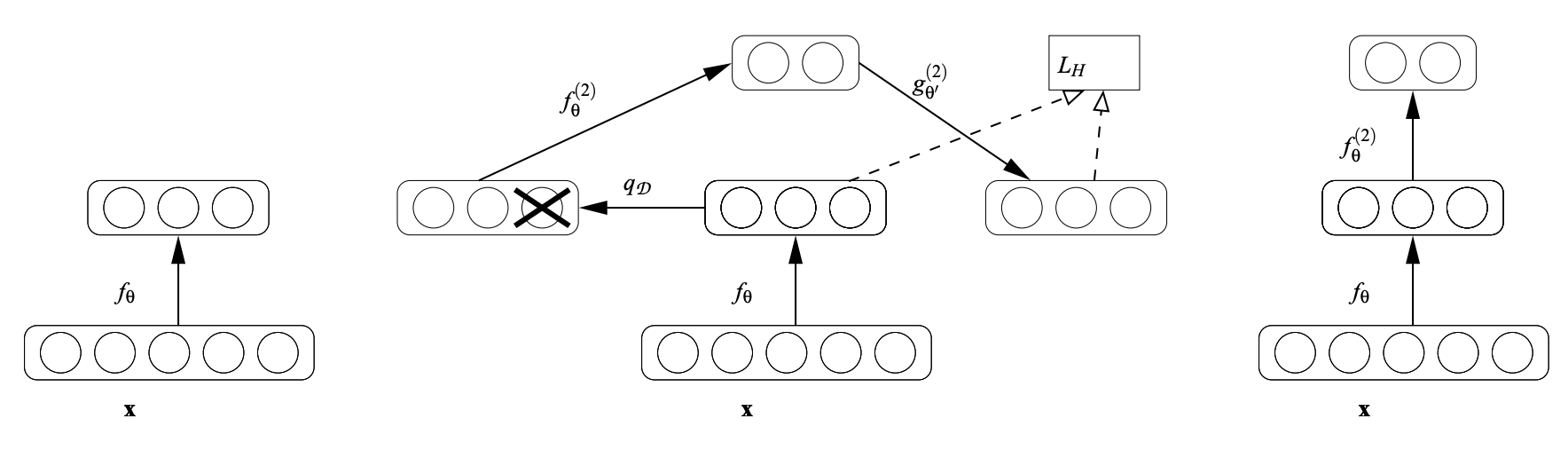}}
    \caption{Left: Autoencoder, Right: Stacking Denoising Autoencoder}
    \label{Img:Fig1}
    \end{figure}

    When the layers in neural networks are very deep, the amount of weight updates is obtained by the multiplication of small gradient descents and may reach zero. Calling the phenomenon of `vanishing gradient', a greedy layerwise training was proposed for this problem, which is a foundation of stacked autoencoders and deep belief networks \cite{hinton2006reducing}. Autoencoder (AE) is one of the unsupervised learning methods and it consists of an encoder $\phi : \mathcal {X} \rightarrow \mathcal {F}$ and a decoder $\psi : \mathcal {F} \rightarrow \mathcal {X}$ which performs as generating high-level or latent value and reconstructing input data (Fig.~\ref{Img:Fig1}). It aims to find $\phi$ and $\psi$, which can make the minimal difference between the given input data and the reconstructed input data ${\phi, \psi = \arg \min_{\phi, \psi}\|X - (\psi \circ \phi) X \|^2 }$ \cite{Goodfellow-et-al-2016}.
    \par AE has been studied for extracting latent characteristic representations and imputing missing data. Several variants exist to the basic model, forcing learned representations of input to assume useful features, which are regularized autoencoders (sparse, denoising, stacked denoising, and contractive autoencoders) and variational autoencoders \cite{wiki:autoencoder, poultney2007efficient, vincent2010stacked, vincent2008extracting, rifai2011contractive}. In particular, sparse autoencoder (SAE) learns representations by allowing only a small number of hidden units to be active and others inactive for sparsity, so that a sparsity penalty encourages the model to learn with some specific areas of the network. Denoising autoencoder (DAE) is trained to reconstruct the corrupted input after the first denoising input, minimizing the same reconstruction loss between the clean input and its reconstruction from hidden representation features. Finally, stacked denoising autoencoder is introduced to make a deep network, like stacking restricted Boltzmann machines (RBMs) in deep belief networks \cite{vincent2010stacked, hinton2006reducing, vincent2008extracting, larochelle2009exploring}, only corrupting input and using the highest level output representation of each autoencoder as another input for the next one to study, which can be found in Fig.~\ref{Img:Fig1}. Unlike traditional autoencoders, variational autoencoders (VAEs) are generative models, like generative adversarial networks (GANs), with encoders that form latent vectors, using mean and standard deviation from sampled inputs and decoders to reconstruct/generate the training data.
    
    \subsection{Deep Belief Networks}
\begin{figure}[!htbp]
    \centering
    \centerline{
    \includegraphics[scale=0.25]{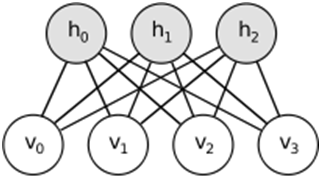}
    \includegraphics[scale=0.25]{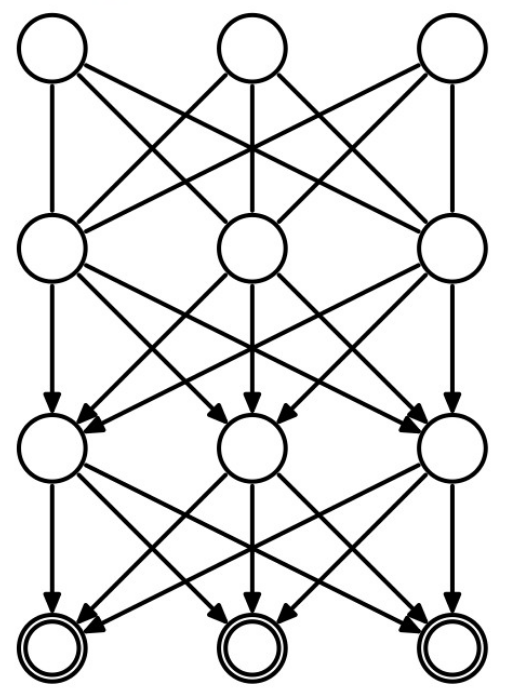}}
    \caption{Left: Restricted Boltzmann Machine with four visible nodes and three hidden nodes, Right: Three-layer Deep Belief Network \cite{Goodfellow-et-al-2016, hinton1986learning, salakhutdinov2009deep, hinton2006fast}}
    \label{Img:Fig2}
    \end{figure}

    Deep belief network (DBN) is composed of a stacked RBM and a belief network \cite{hinton2006fast, hinton1986learning, pearl2014probabilistic}. RBM has a similar concept to AE, but AE has three layers (input, hidden, and output layer) and is deterministic. RBM has two layers (visible and hidden layer) and is stochastic. As pre-training, the first RBM is trained with a sample $v$, a hidden activation vector $h$, a reconstruction $v'$, a resampled hidden activations $h'$ (Gibbs sampling) and weight is updated with $\Delta W = \epsilon (vh^T - v'h'^T)$ (single-step version of contrastive divergence) to get a maximum probability of $v$. The hidden layer represents a new input layer for the second RBM, followed by the first RBM, and the network can start from learning high-level features. When stacked RBMs are all trained, a belief network is added onto the last hidden layer from RBMs and trained to provide a label corresponding to the input label (Fig.~\ref{Img:Fig2}) \cite{hinton1986learning, salakhutdinov2009deep, hinton2006fast}.

    \subsection{Attention Learning}
\begin{figure}[!htbp]
    \centering
    \centerline{
    \includegraphics[scale=0.3]{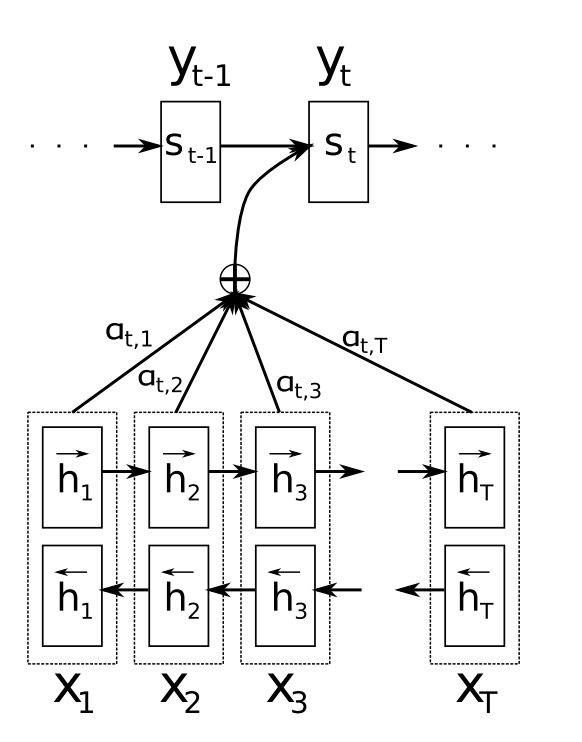}}
    \caption{The encoder-decoder model with additive attention mechanism \cite{bahdanau2014neural}}
    \label{Img:Fig7}
    \end{figure}

    Attention mechanism can be described by mapping a query and a set of key-value pairs to an output. The output is calculated as the weighted sum of the values, and the weight assigned to each value is calculated by the compatibility function of the query with that key \cite{vaswani2017attention}. The mechanism differs in the way of the process, including scaled dot-product attention and multi-head attention. Reference \cite{bahdanau2014neural} introduced neural machine translation (NMT) with attention mechanisms to help memorize long source sentences. The authors proposed a neural machine translation, which consists of an RNN or a bidirectional RNN as an encoder with hidden states and a decoder with a sum of hidden states weighted by alignment scores to emulate searching through a source sentence during decoding a translation (Fig.~\ref{Img:Fig7}).
    \par A RNN with attention has hidden states $h_i$, and a Bi-RNN with attention has forward $\rvec{h}_i$ and backward $\lvec{h}_i$ hidden states (ex. $h_i = [\rvec{h}_i^T; \lvec{h}_i^T]^T$). Each annotation $h_i$ contains information about the whole input sequence with a strong focus on the parts surrounding the i$^t{^h}$ word of the input sequence. The probability $\alpha_i{}_j$ reflects the importance of the annotation $h_j$ with respect to the previous hidden state $s_i{}_-{}_1$ and the context vector $c_i$. The context vector $c_i$ is computed as a weighted sum of the hidden annotations $h_j$. And the weight $\alpha_i{}_j$ of each annotation $h_j$ is computed by how well the inputs around position $j$ and the output at position $i$ match \cite{bahdanau2014neural}. The motivation is for the decoder to decide words to pay attention to. With the context vector which has access to the entire input sequence, rather than forgetting, the alignment between input and output is trained and controlled by the context vector.

    \subsection{Transfer Learning}
    In transfer learning, a base network on a base dataset is first trained, and then a target network is trained on a target dataset using learned features \cite{yosinski2014transferable,caruana1995learning, bengio2012deep, bengio2011deep}. This process is generally meaningful and making a significant improvement when the target dataset is small to train, and researchers intend to avoid overfitting. Usually, after training a base network, the first $n$ layers are copied and used for a target network, and the remaining layers of the target network are randomly initialized. The transferred layers can be left as frozen or fine-tuned, which means either locking the layers so that there is no change during training the target network or backpropagating the errors for both copied and newly initialized layers of the target network \cite{yosinski2014transferable}.

    \subsection{Reinforcement Learning}
    \par
\begin{figure}[!htbp]
    \centering
    \centerline{
    \includegraphics[scale=0.3]{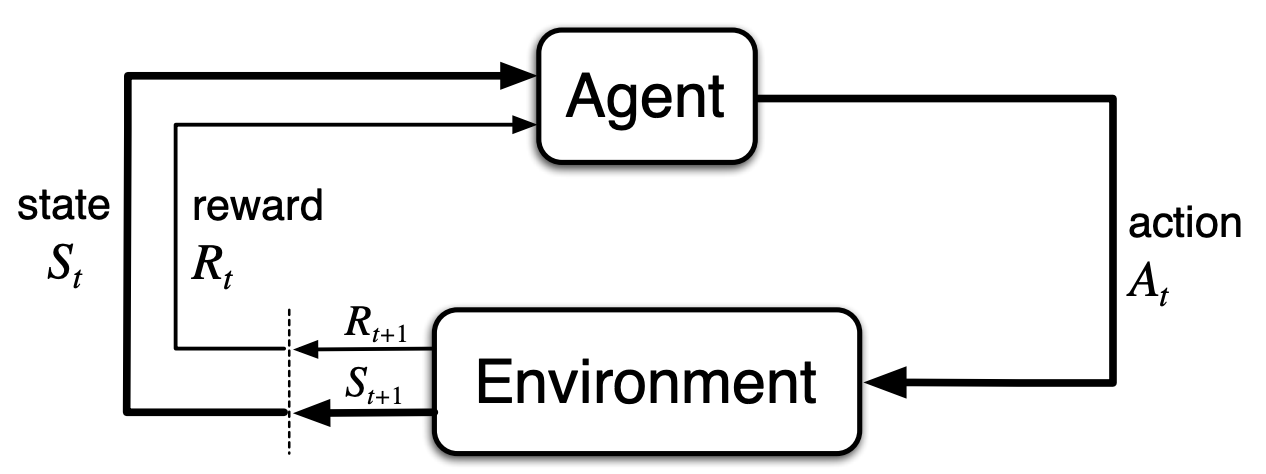}}
    \caption{The agent-environment interaction in reinforcement learning \cite{sutton1998introduction}}
    \label{Img:Fig18}
    \end{figure}

 Reinforcement learning was introduced as an agent learning policy $\pi$ to take action in the environment to maximize cumulative rewards. At each time stamp $t$, an agent observes a state $s_t$ from its environment and takes an action $a_t$ in state $s_t$. The environment and the agent then transition to a new state $s_t{}_+{}_1$ based on the current state $s_t$ and the chosen action, and it provides a scalar reward $r_t{}_+{}_1$ to the agent as feedback \cite{sutton1998introduction} (Fig.~\ref{Img:Fig18}). 
    
    Markov decision process (MDP) is the mathematical formulation of the RL problem. The MDP formulation consists of:
    \begin{itemize}
    \item a set of states $S$
    \item a set of actions $A$
    \item a transition function $P_a(s, s') = Pr(s_t{}_+{}_1 = s' | s_t = s, a_t = a)$\par : from state $s$ to state $s'$ under action $a$ 
    \item a reward after transition $R_a(s, s')$\par : from state $s$ to state $s'$ with action $a$ 
    \item a discount factor $\gamma \in [0,1]$\par : lower values place more emphasis on immediate rewards (ex. $\displaystyle\sum_{t=0}^{T-1} \gamma^t r_t{}_+{}_1$)
    \end{itemize}

    \begin{equation}
      \mathbf \pi^* = \arg \max_\pi \mathbb{E}[R|\pi]
    \end{equation}    
    
    RL's goal is to find the best policy with the maximum expected return, and the RL algorithm class includes value function-based, policy search based and actor-critic based methods using both of the preceding \cite{kaelbling1996reinforcement, doya2002multiple, grondman2012survey}. Deep reinforcement learning (DRL) is based on extending the previous work of RL to higher-dimensional problems. The low-dimensional feature representation and robust function approximation of the neural network allow the DRL to handle the curse of dimensionality and optimize the expected return by a stochastic function \cite{bengio2013representation, heess2015learning, schulman2015gradient, arulkumaran2017brief, sutton1998introduction}.

\section{Applications of artificial intelligence methods}
The use of artificial intelligence for medicine is recent and not thoroughly explored. To estimate the algorithms' trend and performance in health informatics, we searched across several databases with the combination of search terms: (`deep learning' OR `neural network' OR `machine learning' OR `artificial intelligence') and (i) medical imaging (ii) EHR (iii) genomics (iv) sensing and online communication health. Among the articles found, significantly relevant papers regarding each part with applying DL algorithms were briefly reviewed.
  
    \subsection{Medical Imaging}
    The first applications of deep learning on medical datasets were medical images including MRI, PET, CT, X-ray, microscopy, ultrasound (US), mammography (MG), hematoxylin \& eosin histology images (H\&E), optical images, etc. PET scans show regional metabolic information through positron emission, unlike CT and MRI scans, which reveal the structural information of organs or lesions within the body in perspective with radio waves with X-rays and magnets. Although medical imaging technology has been chosen for purposes, in terms of potential health risks to the human body due to X-rays, low-dose CT (LDCT) scans have also been considered instead of normal-dose CT (NDCT) images. Still, they have drawbacks such as image quality and diagnostic performance. Applications included pathological and psychiatric studies with images of the brain, lung, abdomen, heart, breast, etc. 
    
    \par Image classification is still the preferred approach for medical image research by classifying one or several classes for each image, whether the disease present or absent. Its limitations are that algorithms need numerous labeled training samples in good quality, but it is difficult, and these have been addressed by transfer learning and multi-stream learning. To track the disease's progression or thoroughly use 3-dimensional data, a combination of RNN and CNN was also studied. All other aspects of medical image analysis have also got deep learning exceptionally quickly. It can be divided into object detection for disease detection with location, image segmentation for disease detection with labeling pixels, such as pixel, edge, and region-based image segmentation, class imbalance studies, image registration, image generation, image reconstruction, and other tasks. Image registration is transforming one image set into another set of coordinate systems, and an example is registration of brain CT/MRI images or whole-body PET/CT images for tumor localization.

    \begin{figure}[!htbp]
    \centering
    \centerline{
    \includegraphics[scale=0.27]{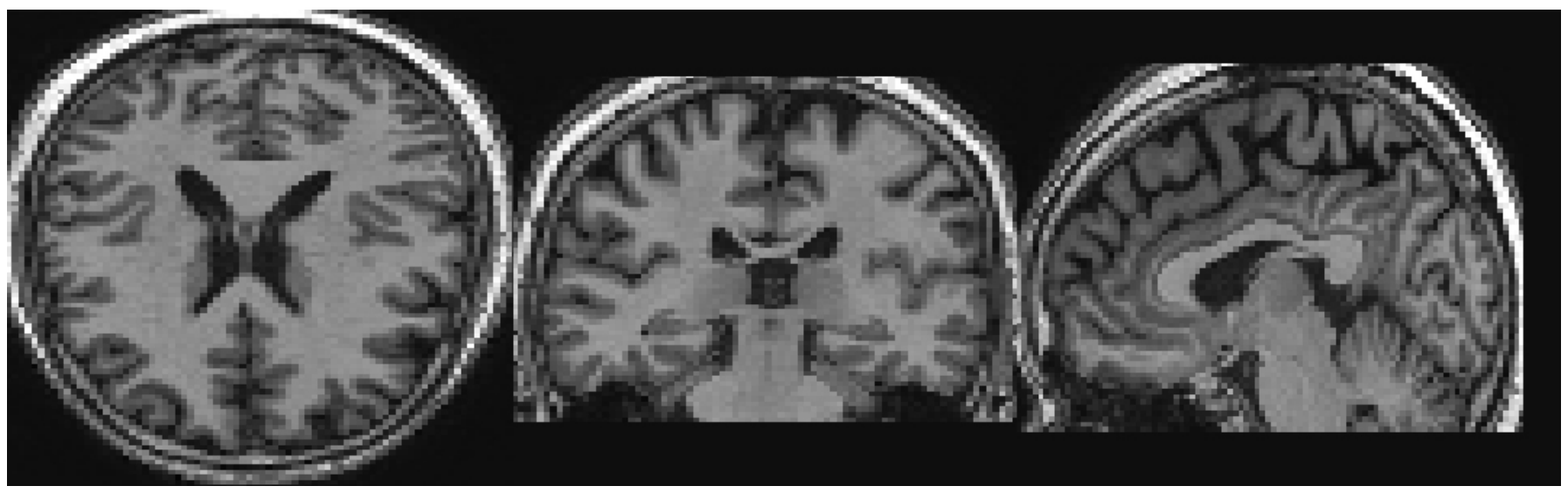}}
    \caption{Slices of an MRI scan of an AD patient, from Left to Right: in axial view, coronal and sagittal views \cite{payan2015predicting}}
    \label{Img:Fig8}
    \end{figure}
       
    \subsubsection{Image Classification and Object Detection}
    \par Since LeNet and AlexNet \cite{lecun1998gradient, krizhevsky2012imagenet}, there has been an exploration of relatively deeper novel architectures widely used as the basis of networks in medical data \cite{szegedy2015going, DBLP:journals/corr/HeZRS15, simonyan2014very, xie2017aggregated, lin2017feature}. Researchers trained the model with or without a pre-trained network. Nevertheless, some of the problems with computer-aided diagnostics (CAD) using medical imaging remain. The challenge is how to use all the features of different shapes and intensities of the detection points, even within the same imaging modality, overlapping detection points, and 3- or 4-dimensional medical images.
    
    \par To deal with this data complexity problem, traditional machine learning with hand-designed feature extraction and deep learning approaches were used \cite{nie20163d,xu2016multimodal,roth2015anatomy,van2016fast,anthimopoulos2016lung,doi:10.1002/jemt.23326,esteva2017dermatologist}. CNN essentially learns the hierarchical structure of complex features to work directly on image patches centered on abnormalities. Disease classification has evolved into 2D/3D CNN, transfer learning through feature extraction with DBN and AE, multi-scale/multi-modality learning, RCNN, and f-CNN. In recent years, a clear transition to deep learning approaches, particularly with transfer learning and multi-stream learning using 3-dimensional images and visual attention mechanisms, has been observed. And the application of these methods does pervasive, from brain MRI to retinal imaging and digital pathology to lung CT.

        \hfill
        \newline
        (1) Transfer Learning
        \par 
        Transfer learning is a popular method in which a model developed for a task is reused as a starting point for a model in other tasks so that researchers do not start the learning process from scratch. Pre-processing with images of similar distribution is still a crucial step influencing classification performance, but performance still has limitations because of a lack of ground-truth/annotated data. The cost and time to collect and manually annotate medical images by experts are enormous, and manual annotation can be even considered subjective. Strategies to alleviate the limitations of the study can be identified in two categories: (i) using pre-trained networks as feature extractors by unsupervised learning-based methods, and (ii) fine-tuning pre-trained networks with natural images or other medical domain images by supervised learning methods. 
        \par For the first category, RBM, DBN, AE, VAE, SAE, and CSAE were unsupervised architectures which constituted a hidden layer with input or visible layers and latent feature representation vectors \cite{hinton2006fast,hinton2006reducing,hinton2012practical,kingma2013auto,vincent2010stacked,vincent2008extracting,larochelle2009exploring, poultney2007efficient, rifai2011contractive}. The medical imaging community also focused on unsupervised learning. After training the layers of unsupervised learning methods, a linear classifier was added to the algorithm's top layer. With a combination of unsupervised learning and classifier (ex. AE with regression, AE with CNN), the methods were applied to the automatic biomarkers extraction and outperformed traditional CAD approaches \cite{brosch2013manifold, plis2014deep, suk2013deep, suk2014hierarchical, van2016combining, cheng2016computer, shan2016deep}. 
        \par Also, to avoid lack of training samples and overfitting, transfer learning via fine-tuning has been proposed in medical imaging applications, using a database of labeled natural images or other labeled medical field images \cite{shin2016deep, chen2015standard, tajbakhsh2016convolutional, xu2016multimodal, samala2016mass, chang2017unsupervised, phan2016transfer, nishio2018computer}. Pre-training supervised learning's layers and copying the first few layers into the new algorithm with the target dataset firstly done, and fine-tuning was performed by optimizing the whole algorithm. There was concern about using natural or other field medical image datasets for fine-tuning since there is a profound difference between those. Nevertheless, previous studies have shown that CNN, fine-tuned based on natural image/other medical field data, improves the performance of algorithms, such as the shape, edges, etc. Even if the base and target datasets are dissimilar, unless the target dataset is significantly smaller than the base dataset, transfer learning is likely to give us a powerful model without overfitting, in general \cite{nishio2018computer, yosinski2014transferable}. For example, in \cite{nishio2018computer}, CNN with or without transfer learning was compared with natural image datasets for classification between benign nodule, primary lung cancer, and metastatic lung cancer, and pre-trained model outperformed others with around 13\% difference of accuracy. Although transfer learning was mostly chosen to combat lack of data with natural images and other domain medical images, recently, \cite{shan20183} proposed a 3D convolutional encoder-decoder network for LDCT images via transfer learning from a 2D trained network. LDCT newly has been used in the medical imaging field because of health risk; however, it makes the low diagnostic performance. The authors introduced a 3D conveying path based convolutional encoder-decoder to denoise LDCT images to normal-dose CT images. As a radiologist scans adjacent slices to extract pathological information more, the authors placed the trained 2D convolutional layer in the middle of the 3D convolutional layers. And they incorporated the 3D spatial information from adjacent image slices.
        
        \hfill
        \newline
        (2) Multi-stream Architectures
        \par 
        Whereas CNN is fundamentally designed for fixed-size and one type of 2D images, medical images are inherently 3- or 4-dimensional images, image sizes are varied, imaging techniques produce different images, and coordinates of detection points are different and comparably small. 
        \par Dimensional problems can be solved using the 3-dimensional image itself. 3D Volume of Interest (VOI) was initially used for classification problems \cite{hosseini2016alzheimer, payan2015predicting}. In general, researchers have developed 3D kernel or convolutional layers and several new layers that formed the basis of their network, and those have shown to outperform existing methods. However, there is a computational burden in processing 3D medical scans. Also, for the voxel size difference, data interpolation was tried but resulted in severely blurred images. Therefore, a dilated convolution and multi-stream learning (multi-scale, multi-angle, multi-modality) were suggested as another solution. 
	
	    \par For multi-stream learning, the default CNN architecture was trained, and the channels were merged at any point in the network. Still, in most cases, the final feature layers were concatenated or summarized to make the final decision on the classifier. Although there have been some studies including 3D Faster-RCNN \cite{zhu2018deeplung} for nodule detection with 3D dual-path blocks and U-Net shaped AE structures, the two most widely used main approaches were multi-scale analysis and 2.5D classification \cite{kamnitsas2017efficient, shen2015multi, moeskops2016automatic, song2015accurate, yang2017cascade, cheng2016active, yang2015automated, kawahara2016multi, nie20163d}. It has become widespread, especially localization, which often requires parsing of 3D volumes and better approaches for classification and segmentation problems, following clinicians' workflow that rotate, zoom in/out 3D images and check adjacent images during diagnosis.
        \par First of all, multi-scale image analysis reliably detected abnormal pixels for irregularly shaped diseases with various intensity distributions and densities \cite{kamnitsas2017efficient, moeskops2016automatic, song2015accurate, shen2015multi, roth2015improving, kawahara2016multi}. For example, \cite{kamnitsas2017efficient} used multi-scale 3D CNN with fully connected CRF for accurate brain lesion segmentation. In \cite{kawahara2016multi}, multi-resolution two-stream CNN was proposed with hybrid pre-trained and skin-lesion trained layers. The authors first trained the original images and each stream with the highest resolution images and low-resolution images created by average pooling. And then, they concatenated the last layers for final prediction. In \cite{shen2015multi} and \cite{ciompi2017towards}, both used multi-scale CNN architecture in multiple streams for pulmonary nodule classification. Reference \cite{shen2015multi} investigated the problem of diagnostic pulmonary nodule classification, which primarily relied on nodule segmentation for regional analysis, and proposed the hierarchical learning framework that used multi-scale CNNs to learn class-specific features without nodule segmentation. In recent studies, researchers used reinforcement learning to improve detection efficiency and performance \cite{ghesu2017multi, alansary2019evaluating, pesce2017learning}. Among them, \cite{ghesu2017multi} proposed a combination of multi-scale and reinforcement learning by reformulating the detection problem as a behavior learning task for an agent in reinforcement learning. The agent was trained to distinguish the target anatomical object in 3D with the optimal navigation path and scale.

    \begin{figure}[!htbp]
    \centering
    \centerline{
    \includegraphics[scale=0.35]{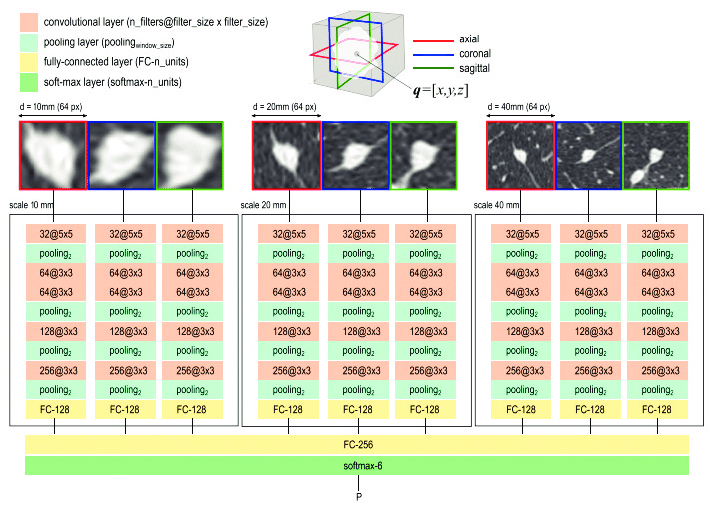}}
    \caption{The multi-angle and multi-scale CNN architecture for pulmonary nodule classification \cite{ciompi2017towards}}
    \label{Img:Fig9}
    \end{figure}
       
        \par 2.5D classification was to address the trade-off between 2D and 3D image classifiers \cite{li2015robust, cheng2016active,yang2015automated, zheng20153d, ciompi2017towards}. The technique used 3D Volume of Interest (VOI), but 2D slices as input images, so it was able to use critical 3D features without compromising computational complexity. It sliced the 3D spatial information in the middle for three or more orthogonal views of an input image or transformed grayscale images into color images. When sliced into three parts based on the intersection of three axial, sagittal and coronal planes, the three 2D slices were generally selected, as shown in Fig.~\ref{Img:Fig9}. Otherwise, image slices were collected variously through scale, random translation, or rotation. For instance, in \cite{ciompi2017towards}, each stream took images of different angles and scales for nine different kinds of perifissural lung nodules classification problems. Without feeding information about nodule size, each stream was trained, followed by a concatenation of the last layers for classifiers, SVM and KNN. 
        
    \begin{figure}[!htbp]
    \centering
    \centerline{
    \includegraphics[scale=0.08]{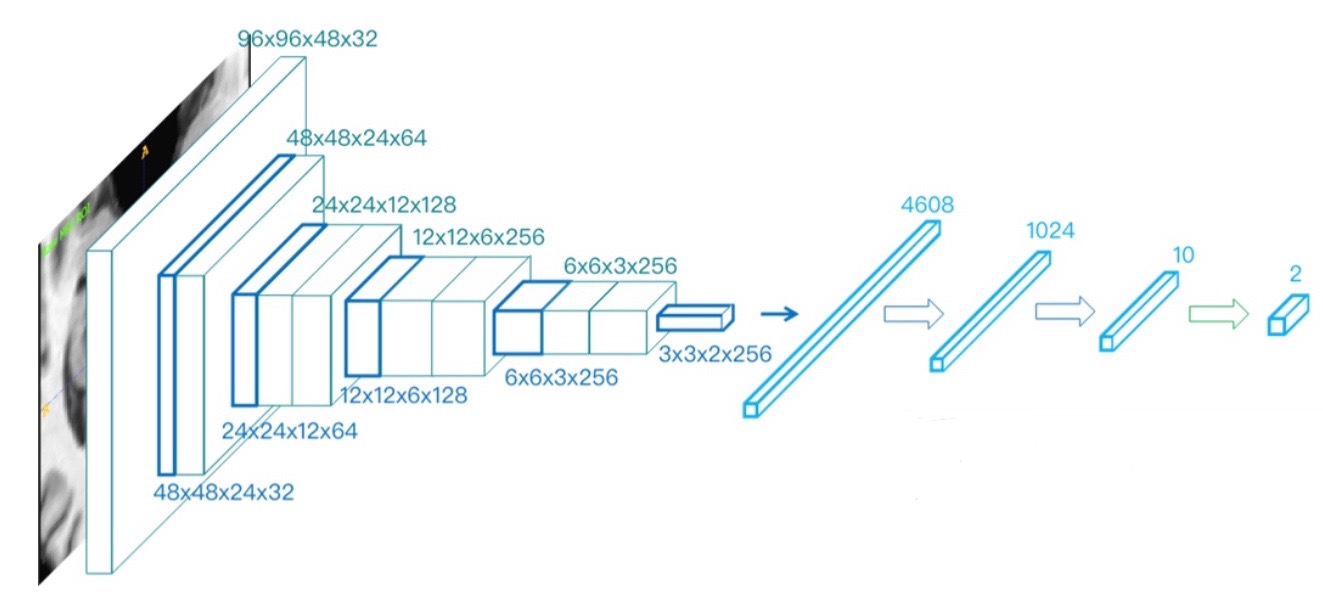}
    \includegraphics[scale=0.11]{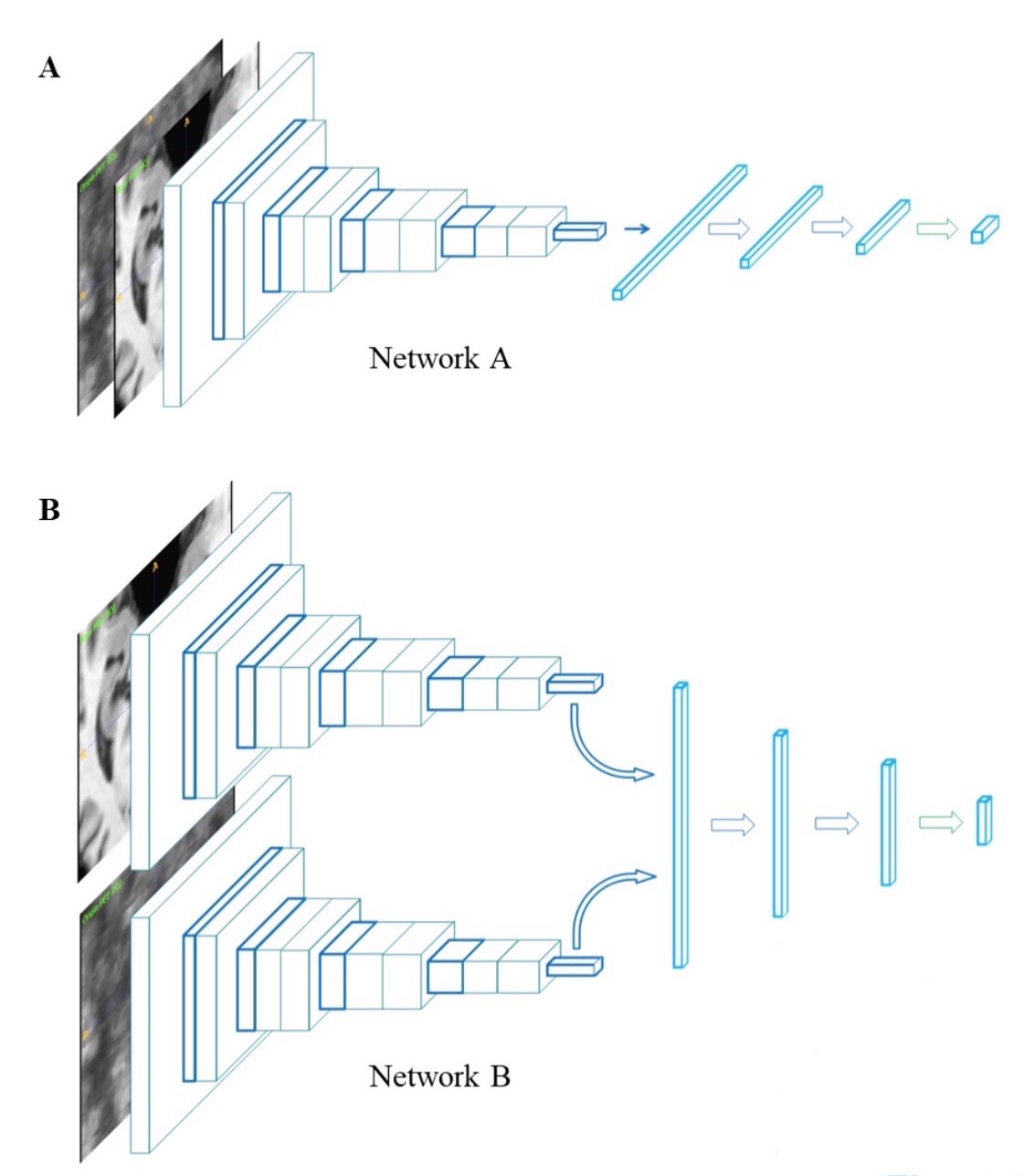}}
    \caption{The architecture of the single- and multi-modality network for Alzheimer's disease \cite{DBLP:journals/corr/abs-1902-09904}}
    \label{Img:Fig10}
    \end{figure}

        \par Third of all, multi-modality was also considered to maximize the usefulness of multiple medical imaging techniques, which have different advantages. In general, PET captures the metabolic information, and CT/MRI does the structural information of organs. As metabolic changes occur before any functional and structural changes in tissues, organs, and bodies, PET facilitates early disease detection \cite{teramoto2016automated, nie20163d,  DBLP:journals/corr/abs-1902-09904}. In \cite{DBLP:journals/corr/abs-1902-09904}, the authors looked into paired FDG-PET and T1-MRI to catch different biomarkers for Alzheimer's disease. PET indicates the regional cerebral metabolic rate of glucose to evaluate tissues' metabolic activity. MRI provides high-resolution structural information of the brain to measure the structural metrics such as thickness, volume, and shape. In particular, the study measured the shrinkage of cerebral cortices (brain atrophy) and hippocampus (memory-related), enlargement of ventricles, and change of regional glucose uptake. Image registration was necessary to use different modality images together. The authors showed the performance and potential of the architecture for multi-modality, comparing single-modality with multi-modality (sharing weights like 3D CNN) and multi-modality (multiple streams without sharing weights) (Fig.~\ref{Img:Fig10}). 

        \subsubsection{Image Segmentation}
        \par Image segmentation is a process of partitioning an image into multiple meaningful segments (sets of pixels) in a bottom-up approach. And CNN-based models are still the most commonly used to classify each pixel in images, in terms of shared weights compared to a fully connected network. Nevertheless, a drawback of this approach is huge overlaps from neighboring pixels and repeated computations of the same convolutions. In order to have a more efficient convolutional layer, the concepts of fully connected layers and convolutional neural networks were combined. A fully convolutional network (fCNN) was proposed to an entire input image in an efficient way rewriting the fully connected layers as convolutions. While `Shift-and-stitch' \cite{long2015fully} was proposed to boost up the performance of fCNN, U-Net, an image segmentation architecture, was proposed for biomedical images \cite{ronneberger2015u}. Inspired by fCNN, \cite{ronneberger2015u} proposed a U-Net architecture with upsampling (upconvolutional layers) and skip-connection, and it secured a better output. 
        \par A similar approach has been studied by some researchers, and there have been a variety of variant algorithms \cite{milletari2016v, cciccek20163d, drozdzal2016importance, badrinarayanan2017segnet}. More specifically, \cite{cciccek20163d} expanded U-Net from 2D to 3D architecture with introducing use cases of (i) semi-annotation and (ii) full-annotation of training sets. Full annotation of the 3D volume is not only difficult to obtain but also leads to rich training. Therefore, the authors focused on generating 3D models to learn image segmentation with only a few annotated 2D slices for training. In \cite{milletari2016v}, the authors proposed a 3D-variant of U-Net architecture, called V-net, performing 3D image segmentation using 3D convolutional layers and dice coefficient optimization. Since it is not uncommon to have a strong imbalance between the number of foreground and background voxels, the authors proposed an objective function based on dice coefficients instead of previous researchers' approaches, re-weighting. Reference \cite{drozdzal2016importance} investigated the use of short and long ResNet-like skip connections, and \cite{badrinarayanan2017segnet} proposed a SegNet, which reused pooling indices of the decoders to perform up-sampling of the low-resolution feature maps. That was one of the most important key elements of the SegNet, gaining high-frequency details and reducing the number of parameters to train in decoders. Still, the architecture upon the U-Net architecture was built with the nearest neighbor interpolation for upsampling, downsampling and squeeze-and-excitation (SE) residual building blocks, multi-scale and 3D convolutional kernels for adjacent images network \cite{zhu2019anatomynet,  hasan2018modified, limx2019segmentation, li2017multi}. 
        \par Although these specific segmentation architectures offered compelling advantages, researchers have also achieved excellent segmentation results by combining RNN, MRF, CRF, RF, dilated convolutions, and others with segmentation algorithms \cite{xie2016spatial, stollenga2015parallel, andermatt2016multi, chen2016combining, kong2016recognizing, alom2019breast}. Reference \cite{Poudel2016RUNet} combined 2D U-Net architecture with GRU to perform 3D segmentation, and \cite{chen2016combining} applied it several times in multiple directions to incorporate bidirectional information from neighbors. And 3D fCNN with summation operation instead of concatenation and 4D fully convolutional structured LSTM were studied, and those outperformed to the 2D U-Net method using RNN \cite{yu2017volumetric, gao2018fully}. Several fCNN methods have also been tried, using graphical models such as MRF and CRF, applied on top of the likelihood map produced by CNN or fCNN \cite{zhu2018adversarial, shakeri2016sub, alansary2016fast, cai2016multi, christ2016automatic, fu2016deepvessel, Gao2016SegmentationLP}. Finally, researchers have shown dilated convolutional layers and attention mechanisms \cite{yu2015multi, chen2017rethinking, wang2019deep, Mishra2018Ultrasoun}. Reference \cite{yu2015multi} and \cite{chen2017rethinking} employed dilated convolution to handle the problem of segmenting objects at multiple scales and systematically aggregate multi-scale contextual information. Reference \cite{wang2019deep} proposed their automatic prostate segmentation in transrectal ultrasound images, using a 3D deep neural network equipped with an attention module. The attention module was utilized to selectively leverage the multilevel features integrated from different layers and refine the features at each layer. In addition, \cite{Mishra2018Ultrasoun} used fCNN with an attention module for the automatic and accurate segmentation of the ultrasound images, which has broken boundaries.

        \subsubsection{Others (Class Imbalance, Image Registration, Image Generation, etc.)}
        \par One of the key challenges in medical imaging technologies is class imbalance since most voxels/pixels in the image are from non-disease class and often have different numbers of images for each disease. Researchers attempted to solve this problem by adapting a loss function \cite{brosch2016deep} and performing data augmentation on positive samples \cite{kamnitsas2017efficient, litjens2016deep,pereira2016brain}, and etc. The loss function was defined as a larger weight for the specificity to make it less sensitive to data imbalance. In addition, scientists often tried to use images from multiple experiments and multiple tomography techniques, but the resolution, orientation, even dimensionality of the dataset was not the same. The researchers made use of algorithms that attempt to find the best image alignment transformation (registration), generation, reconstruction, and combination of image and text reports \cite{wang2016unsupervised, shin2015interleaved, shin2016learning, kisilev2016medical, karpathy2015deep, shan20183, chen2018statistical, liao2017artificial}. Reference \cite{liao2017artificial} presented a 3D medical image registration method along with an agent, trained end-to-end to perform the registration task coupled with attention-driven hierarchical strategy. And \cite{DBLP:journals/corr/abs-1902-09904} paired FDG-PET and T1-MRI for two different biomarkers for Alzheimer's disease with image registration and multi-modality classifiers. In \cite{shan20183} and \cite{chen2018statistical}, the authors tried to overcome the low diagnostic performance from low-dose CT and low-dose Cone-beam CT (CBCT) images. For low-dose CBCT images, they developed a statistical iterative reconstruction (SIR) algorithm using a pre-trained CNN to overcome data deficiency problems, image noise levels, and image resolution issues. For low-dose CT, they proposed segmentation technology for denoising LDCT images and generating NDCT images.

    \subsection{Electronic Health Records}
\begin{figure}[!htbp]
    \centering
    \centerline{
    \includegraphics[scale=0.4]{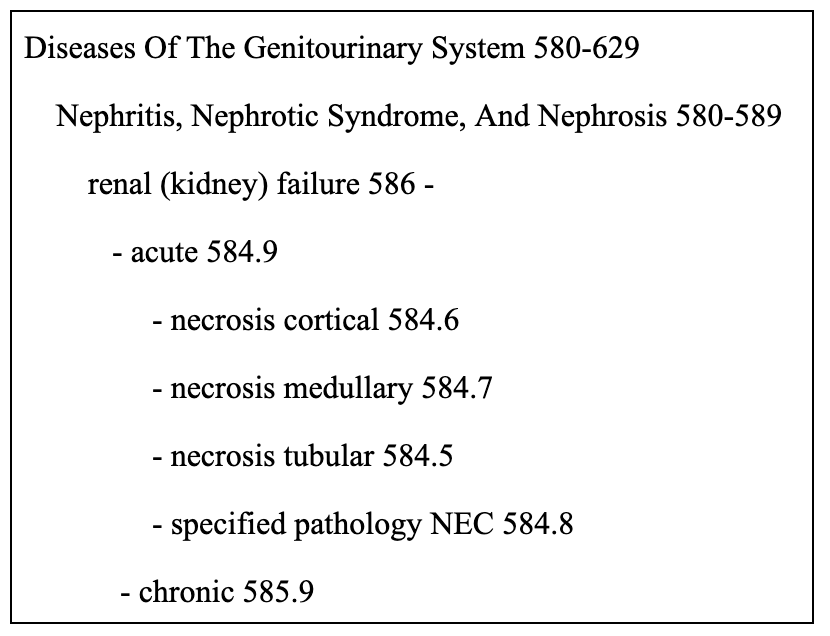}}
    \caption{ICD-9-CM diagnosis codes of acute kidney injury}
    \label{Img:Fig11}
    \end{figure}

    The terms `electronic medical record' and `electronic health record' have been often used interchangeably. EMRs are a digital version of the paper charts in the clinician's office, focusing on the medical and treatment history. And EHRs are designed for sharing the complete health information of patients with other health care providers, such as laboratories and specialists. Since EHR was primarily designed for the internal purpose in the hospital, medical ontologies schema already existed such as the international statistical classification of diseases (ICD) (Fig.~\ref{Img:Fig11}), current procedural terminology (CPT), logical observation identifiers names and codes (LOINC), systematized nomenclature of medicine-clinical terms (SNOMED-CT), unified medical language systems (UMLS) and RxNorm medication code. However, these codes can vary from institution to institution, and even in the same institution, the same clinical phenotype can be represented in different ways across the data \cite{shickel2017deep}. For example, in EHRs, patients diagnosed with acute kidney injury can be identified with a laboratory value of serum creatinine (sCr) level 1.5 times or 0.3 higher than the baseline sCr, presence of 584.9 ICD-9 code, `acute kidney injury' mentioned in the free text clinical notes and so on.
    
    \begin{figure}[!htbp]
    \centering
    \centerline{
    \includegraphics[scale=0.35]{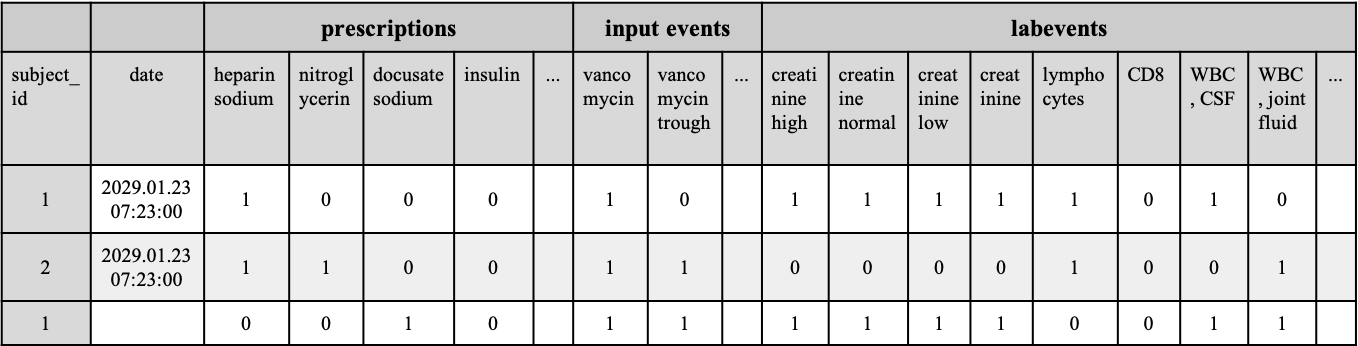}}
    \caption{Sample of pre-processed lab events data; Each row represents a visit to the clinic.}
    \label{Img:Fig12}
    \end{figure}
    
    \par EHR systems include structured data (ex. demographics, diagnostics, physical exams, sensor measurements, vital signs, laboratory tests, prescribed or administered medications, laboratory measurements, observations, fluid balance, procedure codes, diagnostic codes,  hospital length of stay) and unstructured data (ex. notes charted by care providers, imaging reports, observations, survival data) (Fig.~\ref{Img:Fig12}). Challenges in EHR research contain high-dimensionality, heterogeneity, temporal dependency, sparsity and irregularity \cite{hripcsak2012next, jensen2012mining, luo2016big, miotto2017deep, ravi2016deep, lyman2016biomarker, shickel2017deep, esteva2019guide, arXiv190709475L}. EHRs are composed of numerical variables such as 1mg/dl, 5\%, and 5kg, date-time information such as admission time, chart time, date of birth and date of death, and categorical values such as gender, ethnicity, insurance, ICD-9/10 codes (approx. 68,000 codes) and procedure codes (approx. 9,700 codes), and free-text from clinical notes. 
    \par In fact, the data are not only heterogeneous but also very different in distribution. Also, data have discrepancy and bias between patients due to patient transferal, poor alignment of charting events, and bias from retrospective cohorts' data (ex. tendency of treatment frequency depending on the emergency, treatment protocol shift, and different richness of available data focused on certain populations), although it affects models often be ignored \cite{vaso, vaso12, google}. Previous studies have applied artificial intelligence on electronic health records for diseases/admission prediction, information extraction, representation, phenotyping, de-identification, recommendations of medical intervention with or without considering missing data imputations and balancing bias studies.

        \subsubsection{Outcome Prediction}
        \par In studies using deep learning to predict disease, mortality, and readmission from patients' medical records, several studies have shown that one of the main contributions was the characterization of features. Reference \cite{avati2018improving} proposed to improve the palliative care system with a deep learning approach using an observation window and slices. The authors used DNN on EHR data of patients from previous years to predict the mortality of patients within the next 3-12 months period. On the other hand, researchers have increasingly used word embeddings in vectorized representations to predict the outcomes. Reference \cite{choi2016medical} proposed a new method using skip-gram to represent heterogeneous medical concepts (ex. diagnoses, medications, and procedures) based on co-occurrence and predict heart failure with four classifiers (LR, NN, SVM, K-nearest neighbors). Higher-order clinical features may be intuitively meaningful and reduce the dimension of data, but fail to capture inherent information. And raw data may contain all important information, but be represented by a heterogeneous and unstructured mix of elements. Based on the thoughts that related events would occur in a short time difference, the authors applied skip-gram for medical concept vectors and the patient vector with adding the occurred medical vectors to use for heart failure classifier. Using this proposed representation, the area under the ROC curve (AUC) increased by 23\% improvement, compared to the one-hot encoding vector representation. References \cite{nguyen2016mathtt, stojanovic2017modeling,liu2018deep} also treated medical data as language model inputs. Reference \cite{liu2018deep} found that the bag-of-word embedding representing better for their chronic disease prediction case. In \cite{nguyen2016mathtt}, the authors also used CBOW with CNN that captured and exploited the local spatial correlation of the inputs (for input images, pixels are more relevant to closer pixels than faraway pixels). Their system, Deepr, used word2vec (CBOW) and CNN to predict unplanned readmission and motif detection. Not only did they predict discrete clinical event codes as other methods which outperformed than the bag of words and logistic regression models, but they also showed the clinical motif of the convolution filter. Reference \cite{stojanovic2017modeling} generated inpatient representation with both CBOW and skip-gram for each diagnosis and procedure to predict important indicators of healthcare quality (ex. length of stay, total incurred charges, and mortality rates) with regression and classification models. 
        \par There were studies on RNN, LSTM, and GRU for continuous-time signals, including structured data (physical exams, vital signs, laboratory tests, medications) and unstructured data (clinical notes, discharge summary), toward the automatic prediction of diseases and readmission. For example, \cite{pham2016deepcare} and \cite{xiao2018readmission} predicted future risks via deep contextual embedding of clinical concepts. In \cite{pham2016deepcare}, the DeepCare framework used clinical concept word embedding with diagnoses and interventions (medications, procedures). It demonstrated the efficacy of the LSTM based method for disease progression modeling, intervention recommendation, and future risk prediction with time parameterizations to handle irregular timing. In terms of different disease progress rates for each patient, the model was proposed with recency attention (weight) via multi-scale pooling (ex. 12 months, 24 months, all available history). The attention scheme weighted recent events more than old ones. Readmission prediction study was followed by \cite{xiao2018readmission} via contextual embedding of clinical concepts and a hybrid topic recurrent neural network (TopicRNN) model. Emergency room visit prediction was also studied by \cite{Qiao2018ML}, with two non-linear models (XGBoost, RNN) using yearly EHRs. References \cite{esteban2015predicting} and \cite{esteban2016predicting} studied kidney transplantation related complications' prediction with time series based approaches (RNN/LSTM/GRU). They converted static and dynamic features (time-dependent), but in binned formats as low, normal, or high, into latent embedded variables respectively, and then combined. The RNN based models, logistic regression, temporal latent embeddings model, and random prediction models predicted the transplantation three main endpoints: (i) kidney rejection, (ii) kidney loss, and (iii) patient death. Additionally, they pointed out that encoding laboratory measurements were decided to use in a binary way by representing each of them, such as high/normal/low, compared to mean or median imputation and normalization/standardization. In addition, a combination of GRU and the residual network was used to develop a hybrid NN for joint prediction of present and period assertions of medical events in clinical notes \cite{rumeng2017hybrid}. They used the clinical notes (ex. discharge summaries and progress notes), and the prediction outcomes were presence assertions with six categories (ex. present, absent, possible, conditional, hypothetical, and not associated) and the period assertions, including four categories (ex. current, history, future, and unknown). References \cite{vaso33} and \cite{vaso34} also predicted clinical events in intensive care units (ICUs) adopting LSTM and bidirectional LSTM (Bi-LSTM) with an attention mechanism.

        \par To settle the missing value problem, four types of studies were conducted: (i) missing value imputation, (ii) using the percentage of missing values as an input, (iii) using clustering/similarity-based algorithms, and (iv) matching cohorts. Reference \cite{weng2017can} analyzed the percentage of missing values such as demographic details, health status, prescriptions, acute medical outcomes, hospital records, imputed the missing values, and assessed whether machine learning could improve cardiovascular risk prediction with LR, RF, and NN. The cohort of patients is from 30 to 84 years of age at baseline, with complete data on eight core baseline variables (gender, age, smoking status, systolic blood pressure, blood pressure treatment, total cholesterol, HDL cholesterol, diabetes) used in the established ACC/AHA 10-year risk prediction model. Similarly, \cite{che2016interpretable} held experiments on pediatric ICU datasets for Acute Lung Injury (ALI) and proposed a combinatorial architecture of DNN and GRU models in an interpretable mimic learning framework with the imputation process. The DNN was to take static input features, and the GRU model was to take temporal input features. After training a set of 27 static features such as demographic information and admission diagnosis, and another set of 21 temporal features such as monitoring features and discretized scores made by experts with simple imputation method, the authors showed its performance with baseline machine learning methods such as linear SVM, logistic regression (LR), decision tree (DT) and gradient boosting tree (GBT). In recent studies, a hierarchical fuzzy classifier called DFRBS was proposed using a deep rule-based fuzzy classifier and gaussian imputation to predict mortality in intensive care units (ICUs) \cite{davoodi2018mortality}. Reference \cite{golas2018machine} used a clinical text note to show a model for predicting re-hospitalization within 30 days of heart failure patients with interpolation techniques. The risk prediction model was based on the proposed a deep unified network (DUN) model with attention units, a new mesh-like network structure of deep learning designed to avoid over-fitting. Moreover, \cite{che2018recurrent} developed GRU-D network, which was a variation of the recurrent GRU cell for ICD-9 classification and mortality prediction. For missing values, they measured the percentage of them and showed the correlation between this and mortality. With the demonstration of informative missingness, the authors fundamentally addressed the issue with decay rates in GRU-D to directly utilize the missingness with the input feature values and implicitly in the RNN states. Input decay was to use last observation with time passed information, and hidden state decay was to capture more rich knowledge from missingness. Finally, \cite{vaso} pointed out data bias issues that could often be caused by patients' severity and emergencies and approached the problem by imputation and cohort matching. The authors showed the relative data presence rates of vitals signs and laboratory charts that could potentially affect performance. After balancing cohorts' chart presence level and imputing missing values with gaussian process variational autoencoders (GP-VAE), they used the Bi-LSTM in conjunction with the attention mechanism to predict vasopressor needs.

        \subsubsection{Computational Phenotyping}
        \par With the development of electronic health records, researchers retained a large number of patient well-organized datasets. It allows them to reassess existing traditional disease definitions and investigate new definitions and subtypes of diseases more closely. Whereas clinical experts have defined existing diseases along with manuals, but the newly developing computational phenotyping methods aim to find phenotypes and etiologies with data-driven bottom-up approaches. And by new clusters representing novel phenotypes, it is expected to understand the structure and relationships between diseases and provide more beneficial treatment planning with fewer side effects and accompanying complications.
        \par To discover and stratify new phenotypes or subtypes, unsupervised learning approaches, including AE and its variants, have been broadly adopted. For example, \cite{beaulieu2016semi} suggested denoising autoencoders (DAEs) phenotype stratification and random forest (RF) classification. They simulated scenarios of missing and unlabeled data, which is common in EHR, as well as four case/control labeling methods (all cases, one case, percentage case, rule-based case). They randomly corrupted the data and then entered it into the DAE algorithm to extract meaningful features and trained classifiers, including RF with DAE hidden nodes. Through different classifiers and scenarios, the best-generalized algorithm was chosen. In terms of using unlabeled and missing data, they generated those data and conducted trials to see the usefulness of the algorithm in current EHR-based studies. Furthermore, in DeepPatient \cite{miotto2016deep}, a deep neural network consisting of a stack of denoising autoencoders (SDA) was used to capture the stable structure and regular pattern in EHR data representing patients. General demographic details (ex. age, gender, ethnicity), ICD-9 codes, medications, procedures, laboratory tests, and free-text clinical notes were collected and pre-processed, differed by data type, with the open biomedical annotator, SNOMED-CT, UMLS, RxNorm, NegEx, and etc. as other researchers \cite{pivovarov2015learning}. Then, topic modeling and SDA were applied to generalize clinical notes and improve automatic processing. In particular, clinical notes' latent variables were produced with latent dirichlet allocation (LDA) \cite{blei2003latent}, and the frequency of presence of diagnostics, drugs, procedures, and laboratory experiments to extract summarized biomedical concepts and normalized data were calculated. Finally, SDA was used to derive a general-purpose patient representation for clinical predictive modeling, and the performance of DeepPatient was evaluated over disease and patient level. Reference \cite{pivovarov2015learning} also presented the LDA-based unsupervised phenome model (UPhenome), which is a probabilistic graphical model for large-scale discovery of diseases or phenotypes with notes, laboratory tests, medications, and diagnosis codes.

    \begin{figure}[!htbp]
    \centering
    \centerline{
    \includegraphics[scale=0.4]{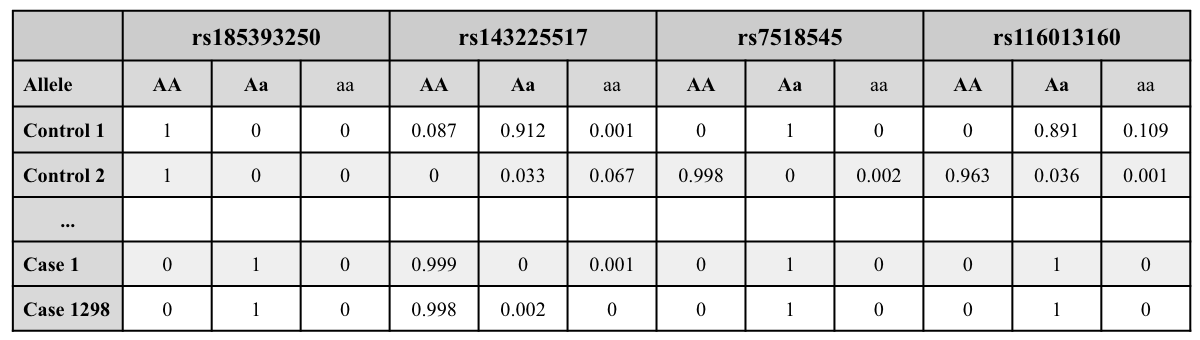}}
    \caption{Sample of the frequency of allele combination; ex. 0.087 for AA (TT) and 0.912 for Aa (TC) or aA (CT). p + q + r = 1 for three alleles}
    \label{Img:Fig13}
    \end{figure}
    
        \par On the other hand, there were computational phenotype studies with other machine learning models, different approaches from AE based, and the association study between genetic variants and phenotypes. Reference \cite{10.1371/journal.pone.0212112} looked into the single nucleotide polymorphism (SNP) rs10455872 which is associated with increased risk of hyperlipidemia and cardiovascular diseases (CVD) and the minor allele frequency (MAF) of the rs10455872 G allele was measured for the SNP (Fig.~\ref{Img:Fig13}). Meanwhile, ICD-9 codes from EHRs were mapped into disease phecodes and the phecodes were used as their inputs for topic modeling \cite{10.1371/journal.pone.0212112,10.1371/journal.pone.0175508}. Topic modeling via non-negative matrix factorization (NMF) was used to extract a set of topics from individuals' phenotype data. The relationship between the topic and LPA SNP was shown with the pearson correlation coefficient (PCC) and LR to determine the most relevant topic for the SNP and the disease. There have been other researches on computational phenotyping to produce clinically interesting phenotypes with matrix/tensor factorization. Also, \cite{Henderson2018Phenotyping} incorporated auxiliary patient information into the phenotype derivation process and introduced their phenotyping through semi-supervised tensor factorization (PSST). In particular, tensors were described with three dimensions (patients, diagnoses, medication), and semi-supervised clustering was proposed using pairs of data points that must be clustered together and pairs that must not be in the same cluster. References \cite{deliu2016identification} and \cite{seymour2019derivation} focused on asthma and sepsis, a kind of heterogeneous disease comprising several subtypes and new phenotypes, caused by different pathophysiologic mechanisms. They stressed that the precise identification of their subtypes and pathophysiological mechanisms with phenotypes might lead clinicians to enable more precise therapeutic and prevention approaches. In particular, both considered non-hierarchical clustering (k-means), but \cite{deliu2016identification} additionally considered hierarchical clustering, latent class analysis, and mixture modeling. Both evaluated the outcomes with phenotype size, clear separation (distance between clusters, soft or hard decision), and characteristics analysis with the distribution. Similar to those, \cite{van2017cluster} suggested log-likelihood and \cite{kyeong2017novel} proposed topological data analysis for subtypes of tinnitus and attention deficit hyperactivity disorder (ADHD). 
        \par Phenotyping algorithms were implemented to identify patients with specific disease phenotypes with EHRs, and the unsupervised based feature selection methods were broadly suggested. However, due to the lack of labeled data, some researchers suggested a fully automated and robust unsupervised feature selection from medical knowledge sources, instead of EHR data. Reference \cite{yu2016surrogate} suggested surrogate-assisted feature extraction (SAFE) for high-throughput phenotyping of coronary artery disease, rheumatoid arthritis, Crohn's disease, and ulcerative colitis, which was typically defined by phenotyping procedure and domain experts. The SAFE contained concept collection, NLP data generation, feature selection, and algorithm training with Elastic-Net. For UMLS concept collection, they used five publicly available knowledge sources, including Wikipedia, Medscape, Merck Manuals Professional Edition, Mayo Clinic Diseases and Conditions, and MedlinePlus Medical Encyclopedia, followed by searching for mentions of candidate concepts. For feature selection, they used majority voting, frequency control, and surrogate selection. The surrogate selection was based on the fact that when S relates to a set of features F only through Y, it is statistically plausible to infer the predictiveness of F for Y based on the predictiveness of F for S. Using low and high threshold for the main NLP and ICD-9 counts, the features were selected and then trained by fitting an adaptive Elastic-Net penalized logistic regression. Also, SEmantics-Driven Feature Extraction (SEDFE) \cite{NING2019103122} showed the performance compared with other algorithms based on EHR for five phenotypes, including coronary artery disease, rheumatoid arthritis, Crohn's disease, ulcerative colitis, and pediatric pulmonary arterial hypertension, and algorithms yielded by SEDFE.
        \par Moreover, there were studies to find new phenotypes and sub-phenotypes and improve current phenotypes by using the supervised learning approach. For example, \cite{cheng2016risk} used a four-layer CNN model with slow temporal fusion (slowly fuses temporal information throughout the network such that higher layers get access to progressively more global information in temporal dimensions) to solve an issue that remained after performing matrix/tensor-based algorithms, extracted phenotypes, and predicted congestive heart failure (CHF) and chronic obstructive pulmonary disease (COPD). References \cite{lipton2015learning} and \cite{che2015deep} framed phenotyping problem as a multilabel classification problem with LSTM and MLP. Reference \cite{che2015deep}'s pre-trained architecture with DAE also showed the usefulness with structured medical ontologies, especially for rare diseases with few training cases. They also developed a novel training procedure to identify key patterns for circulatory disease and septic shock.

        \subsubsection{Knowledge Extraction}
        \par Clinical notes contain dense information about patient status, and information extraction from clinical notes can be a key step towards a semantic understanding of EHRs. It can be started with the sequence labeling or annotation, and Conditional Random Field (CRF) based models have been widely proposed in previous studies. However, researchers newly suggested DNN, and \cite{jagannatha2016structured} was the first group that explored RNN frameworks. EHRs of cancer patients diagnosed with hematological malignancy were used, and the annotated events for notes were broadly divided into two: (i) medication (drug name, dosage, frequency, duration, and route) and (ii) disease (adverse drug events, indication, other sign, symptom or disease), and their RNN based architecture was found to surpass the CRF model significantly. Reference \cite{wu2015named} also showed that DNN outperformed CRFs at the minimal feature setting, achieving the highest F1-score (0.93) to recognize clinical entities in Chinese clinical documents. They developed a deep neural network (DNN) to generate word embeddings from a large unlabeled corpus through unsupervised learning and another DNN for the Named Entity Recognition (NER) task. Unlike word-based maximum likelihood estimation of conditional probability having CRFs, NER used the sentence level log-likelihood approach, which consisted of a convolutional layer, a non-linear layer, and linear layers. On the other hand, \cite{qiu2017deep} implemented CNN to extract ICD-O-3 topographic codes from a corpus of breast and lung cancer pathology reports, using term frequency-inverse document frequency (TF-IDF) as a baseline model. Consistently, CNN outperformed the TF-IDF based classifier. However, not for well-populated classes but for low prevalence classes, pre-training with word embeddings features on differing corpora achieved better performance. In addition, \cite{luo2015subgraph} applied subgraph augmented non-negative tensor factorization (SANTF). That is, the authors converted sentences from clinical notes into a graph representation and then identified important subgraphs. Then the patients were clustered. Simultaneously, latent groups of higher-order features of patient clusters were identified as clinical guidelines, comparing to the widely used non-negative matrix factorization (NMF) and k-means clustering methods. Although several methods of information extraction have already been introduced, \cite{scheurwegs2017assigning} focused on minimal annotation-dependent methods using unsupervised and semi-supervised techniques for multi-word expression extraction, which conveyed a generalizable medical meaning. In particular, the authors used annotated and unannotated corpus of dutch clinical free text. Above that, they proposed a linguistic pattern extraction method based on pointwise linguistic mutual information (LMI), and a bootstrapped pattern mining method (BPM), as introduced by \cite{gupta2014improved}, comparing with a dictionary-based approach (DICT), a majority voting and a bag of words approach. The performance was assessed with a positive impact on diagnostic code prediction. 
        \par Unlike the above, in \cite{Fries2016BrundleflyAS}, the authors extracted time-related medical information from a document collection of clinic and pathology notes from Mayo Clinic with a joint inference-based approach which outperformed RNN. They then found a combination of date canonicalization and distant supervision rules to find time relations with events, using Stanford's DeepDive application \cite{zhang2015deepdive}. DeepDive based system made the best labeling entities to encode domain knowledge and sequence-structure into a probabilistic graphical model. Also, the temporal relationship between an event mention and corresponding document creation time was represented as a classification problem, assigning event attributes from the label set (before, overlap, before/overlap, after).        
        \par As much as it is important to study how medical concepts and temporal events can be explained, relation extraction on medical data including clinical notes, medical papers, Wikipedia, and any other medical-related documents, is also a key step in building medical knowledge graph. Reference \cite{lv2016clinical} proposed a CRF model for a relation classification model and three deep learning models for optimizing extracted contextual features of concepts. Among the three models, deepSAE was chosen, developed for contextual feature optimization with both autoencoder and sparsity limitation remedy solution. They divided the clinic narratives such as discharge summaries or progress notes into complete noun phrases (NPs) and adjective phrases (APs), and relation extraction aimed to determine the type of relationship such as `treatment improves medical problem', `test reveals medical problem', etc. Reference \cite{ling2017diagnostic} extracted clinical concepts from free clinical narratives with relevant external resources (Wikipedia, Mayo Clinic), and trained Deep Q-Network (DQN) with two states (current clinical concepts, candidate concepts from external articles) to optimize the reward function to extract clinical concepts that best describe a correct diagnosis. 
        \par In \cite{li2018extraction}, nine entity types such as medications, indications, and adverse drug events (ADEs) and seven types of relations between these entities are extracted from electronic health record (EHR) notes via natural language processing (NLP). They used the Bi-LSTM conditional random field network to recognize entities and the Bi-LSTM attention network to extract relations. Then they proposed with multi-task learning to improve performance (HardMTL, RegMTL, and LearnMTL for hard parameter sharing, parameter regularization, and task relation learning in multi-task learning, respectively). HardMTL further improved the base model, whereas RegMTL and LearnMTL failed to boost the performance. References \cite{munkhdalai2018clinical} and \cite{zhang2019attention} also showed models for clinical relation identification, especially for long-distance intersentential relations. Reference \cite{munkhdalai2018clinical} exploited SVM, RNN, and attention models for nine named entities (ex. medication, indication, severity, ADE) and seven different types of relations (ex. medication-dosage, medication-ADE, severity-ADE). They showed that the SVM model achieved the best average F1-score outperforming all the RNN variations; however, the Bi-LSTM model with attention achieved the best performance among different RNN models. In \cite{zhang2019attention}, they aimed to recognize relations between medical concepts described in Chinese EMRs to enable the automatic processing of clinical texts, with an attention-based deep residual network (ResNet) model. Although they used EMRs as input data instead of notes for information extraction, the residual network-based model reduced the negative impact of corpus noise to parameter learning. The combination of character position attention mechanism enhanced the identification features from different types of entities. More specifically, the model consisted of a vector representation layer (character embedding pre-trained by  word2vec, position embedding), a convolution layer, and a residual network layer. Of all other methods (SVM, CNN based, LSTM based, Bi-LSTM based, ResNet based models), the model achieved the best performance on F1-score and efficiency when matched with annotations from clinical notes.

        \subsubsection{Representation Learning}
        \par Modern EHR systems contain patient-specific information, including vital signs, medications, laboratory measurements, observations, clinical notes, fluid balance, procedure codes, diagnostic codes, etc. The codes and their hierarchy were initially implemented for internal administrative and billing tasks with their relevant ontologies by clinicians. However, recent deep learning approaches have attempted to project discrete codes into vector space, get inherent similarities between medical concepts, represent patients' status with more details, and do more precise predictive tasks. In general, medical concepts and patient representations have been studied through word embedding and unsupervised learning (temporal characteristics, dimension reduction, and dense latent variables). 
        \par For medical concepts, \cite{choi2016learning} showed embeddings of a wide range of concepts in medicine, including diseases, medications, procedures, and laboratory tests. The three types of medical concept embedding with skip-gram were learned from medical journals, medical claims, and clinical narratives, respectively. The one from medical journals was used as the baseline for their two new medical concept embeddings. They identified medical relatedness and medical conceptual similarity for embeddings and performed comparisons between the embeddings. Reference \cite{choi2016multi} addressed the challenges such as (i) combination of sequential and non-sequential information (visits, medical codes and demographic information), (ii) interpretable representations of RNN, (iii) frequent visits, and proposed Med2Vec based on skip-gram, compared with popular baselines such as original skip-gram, GloVe, and stacked autoencoder. For each visit, they generated the corresponding visit representation with a multi-layer perceptron (MLP), concatenating the demographic information to the visit representation information. Similarly, once such vectors were obtained, clustered diagnoses, procedures, and medications were shown with qualitative analysis. Reference \cite{choi2016medical, choi2016using} also vectorized representation information, but used skip-gram for clinical concepts relying on the sequential ordering of medical codes. In particular, they used heterogeneous medical concepts, including diagnoses, medications, and procedures based on co-occurrence, evaluated whether they were generally well grouped by their corresponding categories, and captured the relations between medications and procedures as well as diagnoses with similarity. With mapping medical concepts to similar concept vectors, they predicted heart failure with four classifiers (ex. LR, NN, SVM, K-nearest neighbors) \cite{choi2016medical}, and an RNN model with a 12- to 18-month observation window \cite{choi2016using}. Reference \cite{henriksson2016ensembles} proposed the approaches with ensembles of randomized trees using skip-gram for representations of clinical events. Meanwhile, \cite{tran2015learning} analyzed patients who had at least one encounter with the hospital services and one risk assessment with their EMR-driven non-negative restricted Boltzmann machines (eNRBM) for suicide risk stratification, using two constraints into model parameters: (i) non-negative coefficients, and (ii) structural smoothness. Their framework has led to medical conceptual representations that facilitate intuitive visualizations, automated phenotypes, and risk stratification.
    	\par Likewise, for patient representations, researchers tried to consider word embedding \cite{nguyen2016mathtt, xiao2018readmission, pham2016deepcare, choi2016doctor, mikolov2013efficient}. Deepr system, suggested by \cite{nguyen2016mathtt}, used word embedding and pre-trained CNN with word2vec (CBOW) to predict unplanned readmission. The authors mainly focused on diagnoses and treatments (which involve clinical procedures and medications). Before applying CNN on a sentence, discrete words were represented as continuous vectors with irregular-time information. For that, one-hot coding and word embedding were held, and a convolutional layer became on top of the word embedding layers. The Deepr system predicted discrete clinical event codes and showed the clinical motif of the convolution filter. Reference \cite{pham2016deepcare} developed their DeepCare framework to predict the next disease stages and unplanned readmission. After the exclusion criteria, there were a total of 243 diagnoses, 773 procedures, and 353 drug codes, embedded in the vector space. They extended a vanilla LSTM by (i) parameterizing time to enable irregular timing, (ii) incorporating interventions to reflect their targeted influence in the course of illness and disease progression, (iii) using multi-scale pooling over time (12 months, 24 months, and all available history), and finally (iv) augmenting a neural network to infer about future outcomes. Doctor AI system \cite{choi2016doctor} utilized sequences of (event, time) pairs occurring in each patient's timeline across multiple admissions as input to a GRU network. Patients' observed clinical events for each timestamp were represented with skip-gram embeddings. And the vectorized patient's information was fed into a pre-trained RNN based model, from which future patient statuses could be modeled and predicted. Reference \cite{xiao2018readmission} predicted readmission via contextual embedding of clinical concepts and a hybrid TopicRNN model. 
        \par Aside from simple vector aggregation with word embedding, it was also possible to directly model the patient information using unsupervised learning approaches. Some unsupervised learning methods were used to get dimensionality reduction or latent representation for patients, especially with words such as ICD-9, CPT, LOINC, NDC, procedure codes, and diagnostic codes. Reference \cite{Zhou2018optimizingAE} analyzed patients' health data, using unsupervised deep learning-based feature learning (DFL) framework to automatically learn compact representations from patient health data for efficient clinical decision making. References \cite{mehrabi2015temporal} and \cite{miotto2016deep} (Deep Patient) used stacked RBM and stacked denoising autoencoder (SDA) trained on each patient's temporal diagnosis codes to produce patient's latent representations over time, respectively. Reference \cite{mehrabi2015temporal} paid special attention to temporal aspects of EHR data, constructing a diagnosis matrix for each patient with distinct diagnosis codes per a given time interval. 
        \par Finally, in \cite{suo2018deep}, similarity learning was proposed for patients' representation and personalized healthcare. With CNN, they captured crucial local information, ranked the similarity, and then did disease prediction and patient clustering for diabetes, obesity, and chronic obstructive pulmonary disease (COPD).

        \subsubsection{De-identification}
        \par EHRs, including clinical notes, contain critical information for medical investigations, and most researchers can only access de-identified records to protect the confidentiality of patients. For example, the Health Insurance Portability and Accountability Act (HIPAA) defines 18 types of protected health information (PHI) that need to be removed in clinical notes. A covered entity should be not individually identifiable for the individual or of relatives, employers, or household members of the individual and all the information such as name, geographic subdivisions, all elements of dates, contact information, social security numbers, IP addresses, medical record numbers, biometric identifiers, health plan beneficiary numbers, full-face photographs and any comparable images, account numbers, any other unique identifying number, characteristic, or code, except for some which are required for re-identification \cite{shickel2017deep, Vincze2014DeidentificationIN}. De-identification leads to information loss, which may limit the usefulness of the resulting health information in certain circumstances. So, it has been desired to cover entities by de-identification strategies that minimize such loss \cite{shickel2017deep}, with manual, cryptographical, and machine learning methods. In addition to human error, the larger EHR, the more practical, efficient, and reliable algorithms to de-identify patients' records are needed.
        \par Reference \cite{li2014identification} applied the hierarchical clustering method based on varying document types (ex. discharge summaries, history and physical reports, and radiology reports) from Vanderbilt University Medical Center (VUMC) and i2b2 2014 de-identification challenge dataset discharge summaries. Instead, \cite{ dernoncourt2017identification} introduced the de-identification system based on artificial neural networks (ANNs), comparing the performance of the system with others, including CRF based models on two datasets: the i2b2 and the MIMIC-III (`Medical Information Mart for Intensive Care') de-identification dataset. Their framework consisted of a Bi-LSTM and the label sequence optimization, utilizing both the token and character embeddings. Recently, three ensemble methods, combining multiple de-identification models trained from deep learning, shallow learning, and rule-based approaches, represented the stacked learning ensemble as more effective than other methods for de-identification processing i2b2 dataset \cite{kim2018ensemble}. And GAN was also considered to show the possibility of de-identifying EHRs with natural language generation \cite{Lee2018NaturalLG}.

        \subsubsection{Medical Intervention Recommendations}
        \par Considering the correlations among the clinical temporal events, understanding the cause and effect of medical treatment, and suggesting the right treatment on time is one of the key goals. Recent studies suggested medical interventions using artificial intelligence, especially with reinforcement learning and graph neural networks \cite{mattieu,gnnattention,finale,finale-Niranjani,iclrgnn-maybe}. The authors in \cite{mattieu,finale,finale-Niranjani} trained reinforcement learning models to learn policies from the clinical protocol of retrospective data in ICU. Reference \cite{finale-Niranjani} showed off-policy reinforcement learning for the management of mechanical ventilation with determining the best action for sedation at a temporal patient state (ex. demographic information, physiological measurements, ventilator information, level of consciousness, current dosages of different sedatives, and the number of prior intubations, etc.). In \cite{mattieu} and \cite{finale}, the authors mainly focused on vasopressor and IV fluid administration for sepsis and hypotension, related to the length of hospital stay, hospital cost, and high level of mortality. They intended to deduce treatment policies that can be clinically interpretable and adaptable for clinicians in ICU. Besides, References \cite{gnnattention} and \cite{iclrgnn-maybe} proposed graph neural network-based models on the co-occurrence of clinical events to learn structural and temporal information. Among them, \cite{gnnattention} added attention mechanism for interpretation.

    \subsection{Genomics}
    \par
\begin{figure}[!htbp]
    \centering
    \centerline{
    \includegraphics[scale=0.045]{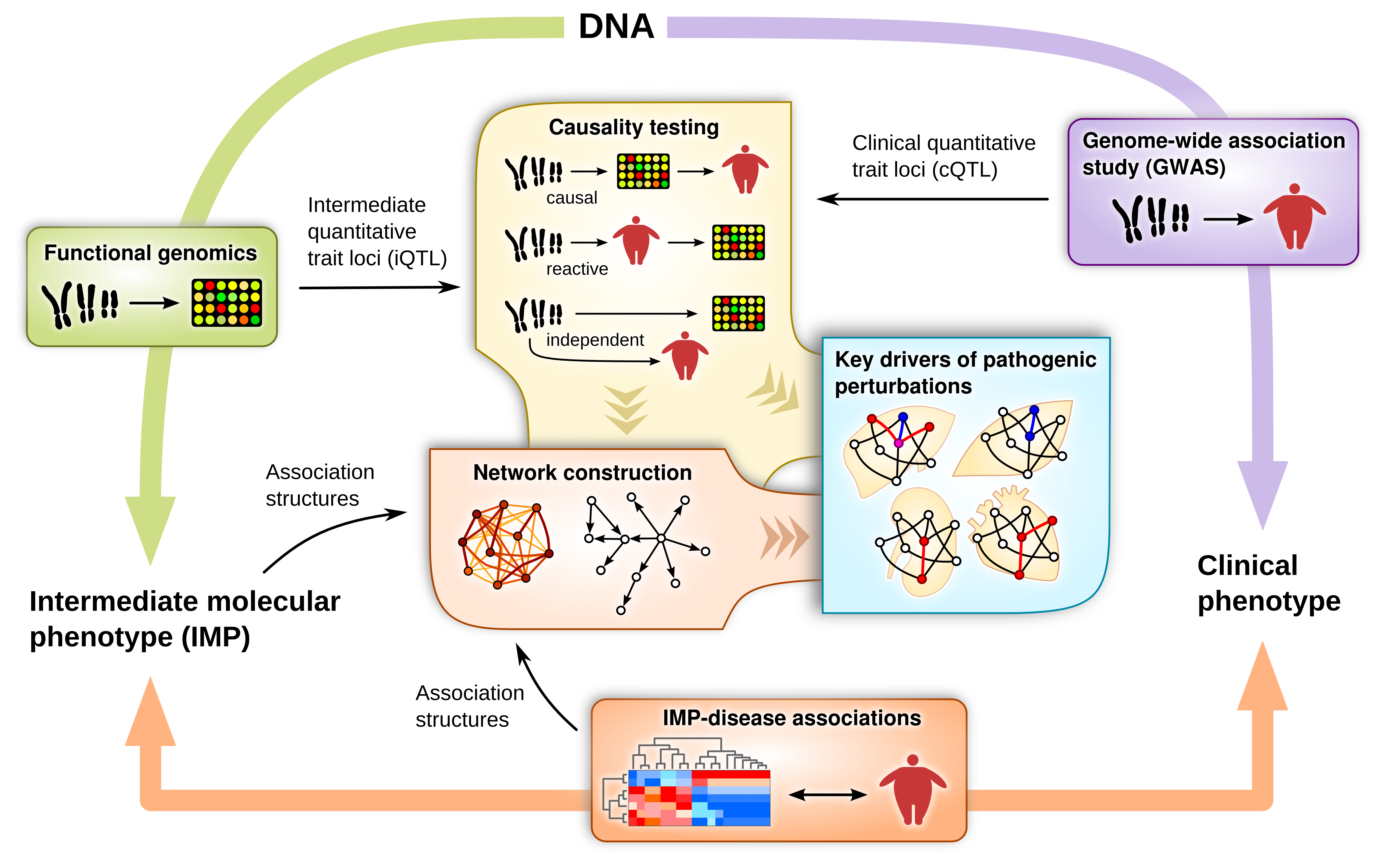}}
    \caption{Systems biology strategies that integrate large-scale genetic, intermediate molecular phenotypes and disease phenotypes \cite{Meng2013Systems}}
    \label{Img:Fig14}
    \end{figure}

 Human genomic data contain vast amounts of data. In general, identifying genes with exploring the function and information structure, investigating how environmental factors affect phenotype, protein formation, interaction without DNA sequence modification, the association between genotype and phenotype, and personalized medicine with different drug responses have been aimed to study \cite{Diao2018Biomedical,angermueller2016deep, gawehn2016deep, eraslan2019deep,pastur2016deep}. More specifically, DNA sequences are collected via microarray or next-generation sequencing (NGS) for specific SNPs only based on a candidate or total sequence as desired. To understand the gene itself after the extraction of the genetic data, what types of mutations can be performed in replication and splicing in transcription have been studied. It is because some mutations and alternative splicings can cause humans to have different sequences and be associated with the disease. Indeed, the absence of the SMN1 gene for infants has been shown to be associated with spinal muscle atrophy and mortality in North America \cite{cartegni2002disruption}. In addition, the environmental factor does not change genotype but phenotype, such as DNA methylation or histone modification, and both genotype and phenotype data can be used to understand human biological processes and disclose environmental effects. Furthermore, it is expected to use analysis to enable disease diagnosis and design of targeted therapies \cite{lyman2016biomarker, collins2015new, gawehn2016deep,eraslan2019deep, leung2015machine, Meng2013Systems}(Fig.~\ref{Img:Fig14}). 
    \par The genetic datasets are incredibly high dimensional, heterogeneous, and unbalanced. As a result, domain experts' pre-processing and feature extraction were frequently required, and recently, machine learning and deep learning approaches were tried to solve the issues \cite{zhang2015deep,kearnes2016molecular}. In terms of feature selection and identifying genes, deep learning helped researchers to capture non-linear features. 

        \subsubsection{Gene Identification}
        \par
\begin{figure}[!htbp]
    \centering
    \centerline{
    \includegraphics[scale=0.23]{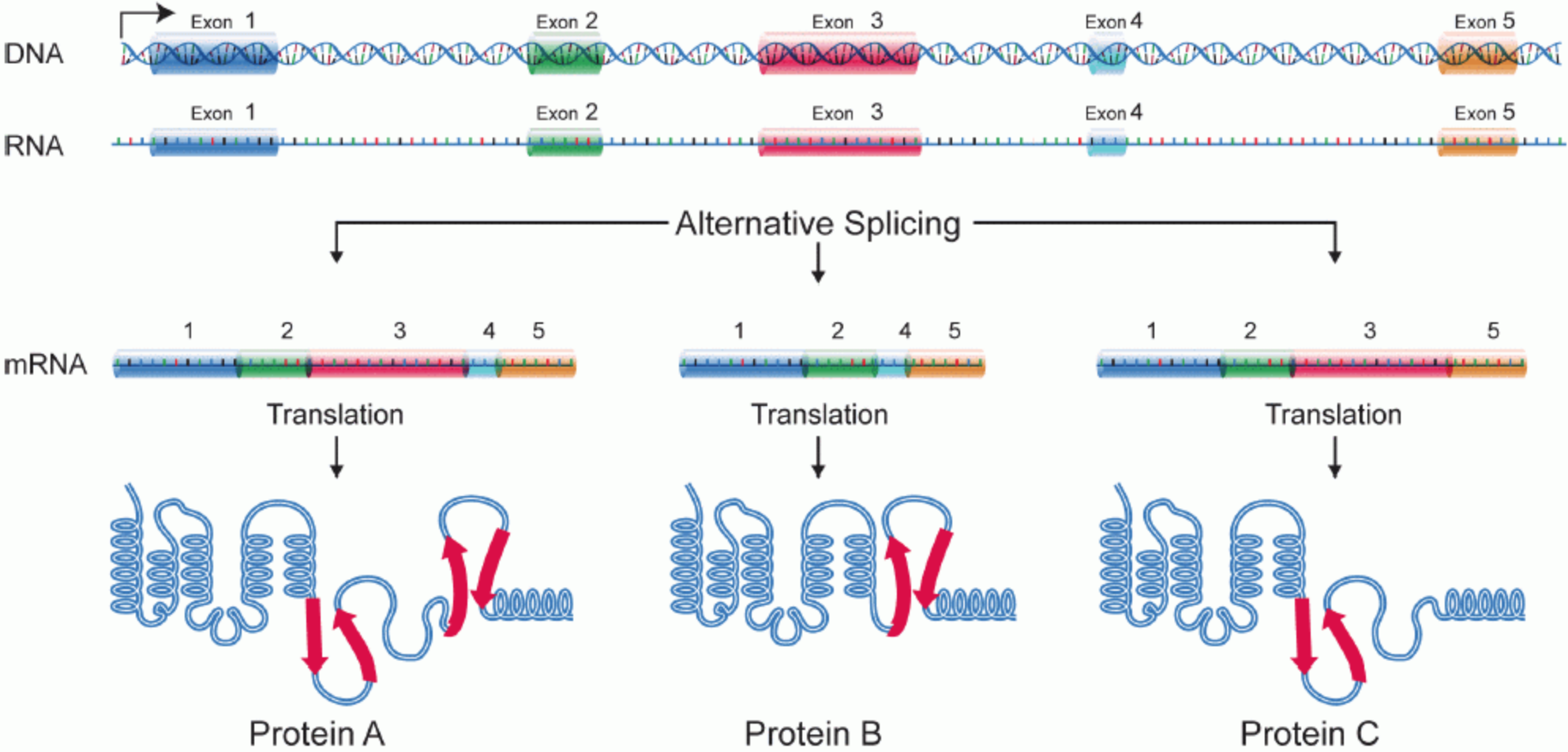}}
    \caption{Alternative splicing produces three protein isoforms \cite{wiki:splicing}.}
    \label{Img:Fig15}
    \end{figure}

 Genomics involves DNA sequencing exploration of the function and information structure of genes. It leads researchers to understand the creation of protein sequences and the association between genotype and phenotype. Analysis of gene or allele identification can help diagnose disease and design targeted therapies \cite{leung2015machine}. And mutations in replication and splicings in transcription were studied to understand the gene itself. For example, DNA mutation is an alteration in the genome's nucleotide sequence, which may and may not produce phenotypic changes in an organism. It can be caused by risk factors such as errors or radiation during DNA replication. Gene splicing is a form of post-transcriptional modification processing, of which alternative splicing is the splicing of a single gene into multiple proteins. During RNA splicing, introns (non-coding) and exons (coding) are split, and exons are joined together to be transformed into an mRNA. Meanwhile, abnormal splicing can happen, including skipping exons, joining introns, duplicating exons, back-splicing, etc. as shown in Fig.~\ref{Img:Fig15}. Predicting mutations and splicing code patterns and identifying genetic variations were chosen, because they are critical for shaping the basis of clinical judgment and classifying diseases \cite{quang2014dann, Li2019Xrare, ibrahim2014multi, chaabane2019circdeep, zeng2016convolutional, tatomer2017unchartered, min2017chromatin, hu2018deephint, lanchantin2016deep,wang2019deep}. 
        
        \par Reference \cite{quang2014dann} proposed a DNN-based model and compared the performance of the traditional models. The traditional combined annotation-dependent depletion (CANN) annotated both coding and non-coding variants, and the authors trained SVM to separate observed genetic variants from simulated genetic variants. With DNN based DANN, they focused on capturing non-linear relationships among the features and reduced the error rate. In \cite{chaabane2019circdeep} and \cite{tatomer2017unchartered}, the authors particularly studied circular RNA, produced through back-splicing and one of the focus of scientific studies due to its association with various diseases, including cancer. Reference \cite{chaabane2019circdeep} distinguished non-coding RNAs from protein-coding gene transcripts, and separated short and long non-coding RNAs to predict circular RNAs from other long non-coding RNAs (lncRNAs). They proposed ACNN-BLSTM, which used an asymmetric convolutional neural network that described sequences using k-mer and a sliding window approach and then Bi-LSTM to describe the sequences, compared to other CNN and LSTM based architectures.
        \par To understand the cause and phenotype of the disease, unsupervised learning, active learning, reinforcement learning, attention mechanisms, etc. were used. For example, to diagnose the genetic cause of rare Mendelian diseases, \cite{Li2019Xrare} proposed a highly tolerant phenotype similarity scoring method of noise and imprecision in clinical phenotypes, using the amount of information content concepts from phenotype terms. In \cite{10.1371/journal.pone.0212112}, the authors identified relationships between phenotypes that result from topic modeling with EHRs and the minor allele frequency (MAF) of the single nucleotide polymorphism (SNP) rs104455872 in lipoprotein which is associated with increased risk of hyperlipidemia and cardiovascular disease (CVD). Reference \cite{baker2016reward} focused on the development and expression of the midbrain dopamine system since dopamine-related genes are partially responsible for vulnerability to addiction. They adopted a reinforcement learning-based method to bridge the gap between genes the behaviour in drug addiction and found a relationship between the DRD4-521T dopamine receptor genotype and substance misuse. 
        \par Meanwhile, AE based methods were applied to generalize meaningful and important properties of the input distribution across all input samples. In \cite{danaee2017deep}, the authors adopted SDAE to detect functional features and capture key biological principles and highly interactive genes in breast cancer data. Also, in \cite{sharifi2019deep}, to predict prostate cancer, two DAEs with transfer learning for labeled and unlabeled dataset's feature extraction were introduced. To capture information for both labeled and unlabeled data, the authors trained two DAEs separately and apply transfer learning to bridge the gap between them. In addition, \cite{ibrahim2014multi} proposed a DBN with an active learning approach to find the most discriminative genes/miRNAs to enhance disease classifiers and to mitigate the dimensionality curse problem. Considering group features instead of individual ones, they showed the data representation in multiple levels of abstraction, allowing for better discrimination between different classes. Their method outperformed classical feature selection methods in hepatocellular carcinoma, lung cancer, and breast cancer. Moreover, \cite{hu2018deephint} developed an attention-based CNN framework for human immunodeficiency virus type 1 (HIV-1) genome integration with DNA sequences with or without epigenetic information. Their framework accurately predicted known HIV integration sites in the HEK293T cell line, and the attention-based learning allowed them to make which 8-bp sequences are important to predict sites. And they also calculated the enrichment of binding motifs of known mammalian DNA binding proteins to exploit important sequences further. In addition to transcription factors prediction, motif extraction strategies were also studied \cite{lanchantin2016deep}. 

        \subsubsection{Epigenomics}
        \par Epigenomics aims to investigate the epigenetic modifications on the genetic material such as DNA or histones of a cell, which affect gene expression without altering the DNA sequence itself. Understanding how environmental factors and higher-level processes affect phenotypes and protein formation and predicting their interactions such as protein-protein or compound-protein interactions on structural molecular information are important. It is because those are expected to perform virtual screening for drug discovery so that researchers can find possible toxic substances and provide a way for how certain drugs can influence certain cells. DNA methylation and histone modification are some of the best-characterized epigenetic processes. DNA methylation is the process by which methyl groups are added to a DNA molecule, altering gene expression without changing the sequence. Also, histones do not affect sequence changes but affect the phenotype. According to the development of biotechnology, it became even much further possible to reduce the cost of collecting genome sequencing and analyze the processes. 
        \par In previous studies, DNN was used to predict DNA methylation states from DNA sequence and incomplete methylation profiles in single cells, and they provided insights with the parameters into the effect of sequence composition on methylation variability \cite{angermueller2016accurate}. Likewise, in \cite{alipanahi2015predicting}, CNN was applied to predict specificities of DNA- and RNA-binding proteins, chromatin marks from DNA sequence and DNA methylation states, and \cite{koh2017denoising} applied a convolutional denoising algorithm to learn a mapping from suboptimal to high-quality histone ChIP-sequencing data, which identifies the binding with chromatin immunoprecipitation (ChIP). 
        \par While DNN and CNN were the most widely used architectures for extracting features from DNA sequences, other unsupervised approaches also have been proposed. In \cite{firouzi2018unsupervised}, clustering was introduced to identify single-gene (Mendelian) disease as well as autism subtypes and discern signatures. They conducted pre-filtration for most promising methylation sites, iterated clustering the features to identify co-varying sites to further refine the signatures to build an effective clustering framework. Meanwhile, in view of the fact that RNA-binding proteins (RBPs) are important in the post-transcriptional modification, \cite{zhang2015deep} developed the multi-modal DBN framework to identify RNA-binding proteins (RBPs)' preferences and predict the binding sites of them. The multi-modal DBN was modeled with the joint distribution of the RNA base sequence and structural profiles together with its label (1D, 2D, 3D, label), to predict candidate binding sites and discover potential binding motifs. 

        \subsubsection{Drug Design}
        \par Identification of genes opens the era for researchers to enable the design of targeted therapies. Individuals may react differently to the same drug, and drugs that are expected to attack the source of the disease may result in limited metabolic and toxic restriction. Therefore, an individual's drug response by differences in genes has been studied to design more personalized treatment drugs while reducing side effects and also develop the virtual screening by training supervised classifiers to predict interactions between targets and small molecules \cite{leung2015machine, ramsundar2015massively}. 
        \par Reference \cite{kearnes2016molecular} showed the structural information, considering a molecular structure and its bonds as a graph and edges. Although their graph convolution model did not outperform all fingerprint-based methods, they represented a new potential research paradigm in drug discovery. Meanwhile, \cite{segler2017generating} and \cite{yuan2017chemical} reported their studies using RNNs to generate novel chemical structures. In \cite{segler2017generating}, they employed the SMILES format to get sequences of single letters, strings, or words. Using SMILES, molecular graphs were described compactly as human-readable strings (ex. c1ccccc), to input strings as an input and get strings as an output from pre-trained three stacked LSTM layers. The model was first trained on a large dataset for a general set of molecules, then retrained on the smaller dataset of specific molecules. In \cite{yuan2017chemical}, their library generation method was described, machine-based identification of molecules inside characterized space (MIMICS) to apply the methodology toward drug design applications. 
        \par On the other hand, there were studies of other deep learning-based methods in chemoinformatics. In \cite{gomez2018automatic, kadurin2017drugan, blaschke2018application, dincer2018deepprofile}, VAE based models were applied, and among them, \cite{gomez2018automatic} generated mapping chemical structures (SMILES strings) into latent space, and then used the latent vector to represent the molecular structure and transformed again in SMILES format. In \cite{dincer2018deepprofile}, their framework was presented to extract latent variables for phenotype extraction using VAEs to predict response to hundreds of cancer drugs based on gene expression data for acute myeloid leukemia. Also, \cite{jaques2017sequence} applied reinforcement learning with RNN to generate high yields of synthetically accessible drug-like molecules with SMILES characters. In \cite{ramsundar2015massively}, the authors examined several aspects of the multi-task framework to achieve effective virtual screening. They demonstrated that more multi-task networks improve performance over single-task models, and the total amount of data contributes significantly to the multi-task effect.

    \subsection{Sensing and Online Communication Health}
    \par Biosensors are wearable, implantable, and ambient devices that convert biological responses into electro-optical signals and make continuous monitoring of health and well-being possible, even with a variety of mobile apps. Since EHRs often lack patients' self-reported experiences, human activities, and vital signs outside of clinical settings, continuously tracking those is expected to improve treatment outcomes by closely analyzing patient's condition \cite{johnson2016machine, yin2017power, yin2019systematic, amengual2018status, ulate2016automated}. Online environments, including social media platforms and online health communities, are expected to help individuals share information, know their health status, and also provide a new era of precision medicine as well as infectious diseases and health policies \cite{zhang2017online, zhang2017longitudinal, zhang2017cataloguing, ma2014knowledge, perrin2015social, Pittman2016social, cookingham2015impact,covidment1,covidment2}.
        
        \subsubsection{Sensing}
        \par
\begin{figure}[!htbp]
    \centering
    \centerline{
    \includegraphics[scale=0.3]{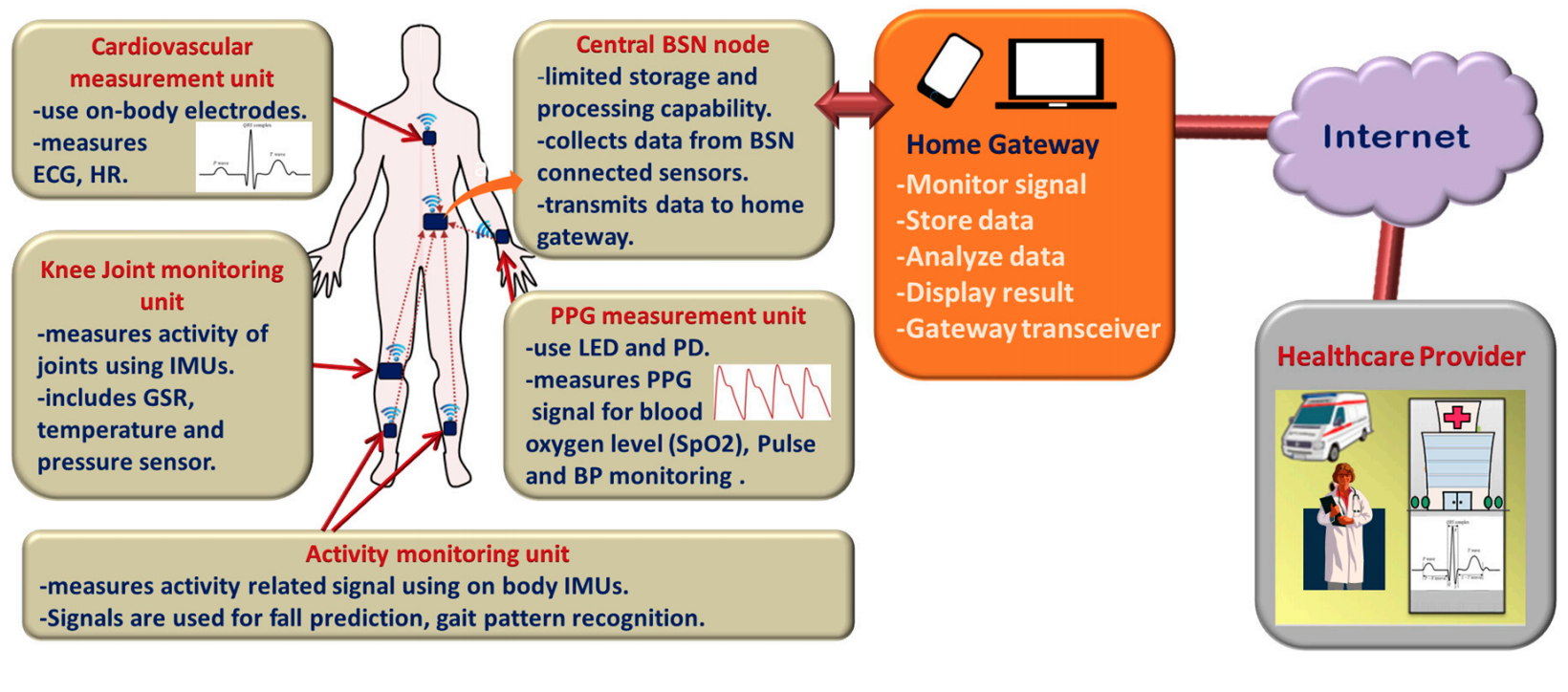}}
    \caption{Schematic overview of the remote health monitoring system with the most used possible sensors worn on different locations (ex. the chest, legs, or fingers) \cite{majumder2017wearable}}
    \label{Img:Fig16}
    \end{figure}

 For decades, various types of sensors have been used for signal recording, abnormal signal detection, and more recent predictions. With the development of feasible, effective, and accurate wearable devices, electronic health (eHealth) and mobile health (mHealth) applications and telemonitoring concepts have also recently been incorporated into patient care \cite{majumder2017wearable, trung2016flexible, Patel2012review, cao2016improving, Yablowitz2018Areview}(Fig.~\ref{Img:Fig16}).
        
        \par Especially for elderly patients with chronic diseases and critical care, biosensors can be utilized to track vital signs, such as blood pressure, respiration rate, and body temperature. It can detect abnormalities in vital signs to anticipate extreme health status in advance and provide health information before hospital admission. Even though continuous signals (ex. EEG, ECG, EMG, etc.) vary from patient to patient and are difficult to control due to noise and artifacts, deep learning approaches have been proposed to solve the problems. In addition, emergency intervention apps were also being developed to speed the arrival of relevant treatments \cite{Yablowitz2018Areview, Patel2012review}.
        \par For example, \cite{Iqbal2018Deep} pointed out that some cardiac diseases, such as myocardial infarction (MI) and atrial fibrillation (Af), require special attention, and classified MI and Af with three steps of the deep deterministic learning (DDL). First, they detected an R peak based on fixed threshold values and extracted time-domain features. The extracted features were used to recognize patterns and divided into three classes with ANN and finally executed to detect MI and Af. Reference \cite{munir2019fusead} presented an anomaly detection technique, Fuse AD, with streaming data. The first step was forecasting models for the next time-stamp with autoregressive integrated moving average (ARIMA) and CNN for a given time-series data. The predicted results were fed into the anomaly detector module to detect whether each time-stamp was normal or abnormal.
        \par To monitor and detect Parkinson's disease symptoms (mainly tremor, freezing of gait, bradykinesia, and dyskinesia), accelerometer, gyroscope, cortical and subcortical recordings with the mobile application has been used. Parkinson's disease (PD) arises with the death of neurons that produce dopamine controlling the body's movement. Hence, to detect the brain abnormality and early diagnose PD, 14 channels from EEG were used in \cite{oh2018deep} with CNN. As neurons die, the amount of dopamine produced in the brain is reduced, and different patterns are created for each channel to classify PD patients. In particular, \cite{eskofier2016recent} specifically focused on the detection of bradykinesia with CNN. They made 5 seconds non-overlapping segments from sensor data for each patient and used eight standard features, widely used as a standard set (total signal energy, the maximum, minimum, mean, variance, skewness, kurtosis, frequency content of signals) \cite{patel2009monitoring, barth2011biometric}. After normalization and training classification, CNN outperformed any others in terms of classification rate. For cardiac arrhythmia detection, CNN based approach was also used by \cite{yildirim2018arrhythmia}, based on long term ECG signal analysis with long-duration raw ECG signals. They used 10 seconds segments and trained the classifier for 13, 15, and 17 cardiac arrhythmia diagnostic classes. Reference \cite{Li2019DeepBN} used pre-trained DBN to classify spatial hyperspectral sensor data, with logistic regression as a classifier. References \cite{amengual2018status} and \cite{ulate2016automated} also covered the potentiality of automatic seizure detection with detection modalities such as accelerometer, gyroscope, electrodermal activity, mattress sensors, surface electromyography, video detection systems, and peripheral temperature. 
        \par Obesity has been identified as one of the growing epidemic health problems and has been linked to many chronic diseases such as type 2 diabetes and cardiovascular disease. Smartphone-based systems and wearable devices have been proposed to control calorie intake and emissions \cite{pouladzadeh2016food, kuhad2015using, mezgec2017nutrinet, hochberg2016encouraging}. For instance, a deep convolutional neural network architecture, called NutriNet, was proposed \cite{mezgec2017nutrinet}. They achieved a classification accuracy of 86.72\%, along with an accuracy of 94.47\% on a detection dataset, and the algorithm also performed a real-world test on datasets of self-acquired images, combined with pictures from Parkinson's disease patients, all taken using a smartphone camera. This model was expected to be used in the form of a mobile app for the Parkinson's disease dietary assessment, so it was essential to enable real situations for practical use. In addition, mobile health technologies for resource-poor and marginalized communities were also studied with reading X-ray images taken by a mobile phone \cite{cao2016improving}.

        \subsubsection{Online Communication Health}
        \par Based on online data that patients or their parents wrote about symptoms, there were studies that helped individuals, including pain, fatigue, sleep, weight changes, emotions, feelings, drugs, and nutrition \cite{opitz2014breast,wilson2014finding,de2014mental,de2013predicting,marshall2016symptom,ping2016breast,yang2016mining,de2017gender,zhang2017longitudinal,covidment1,covidment2}. For mental issues, writing and linguistic style and posting frequency were important to analyze symptoms and predict outcomes. 
        \par Suicide is among the ten most common causes of death, as assessed by the World Health Organization. References \cite{kumar2015detecting} and \cite{coppersmith2018natural} pointed out that social media can offer new types of data to understand the behavior and pervasiveness and prevent any attempts and serial suicides. Both detected quantifiable signals around suicide attempts and how people are affected by celebrity suicides with natural language processing. Reference \cite{kumar2015detecting} used n-gram with topic modeling. The contents before and after celebrity suicide were analyzed, focusing on negative emotional expressions. Topic modeling by latent dirichlet allocation (LDA) was held on posts shared during two weeks preceding and succeeding the celebrity suicide events to measure topic increases in post-suicide periods. In \cite{coppersmith2018natural}, they initialized the model with pre-trained GloVe embeddings, and word vectors sequences were processed via a bi-LSTM layer with using skip connections into a self-attention layer, to capture contextual information between words and apply weights to the most informative subsequences. Meanwhile, in the fact that patients tend to discuss the diagnosis at an early stage \cite{zhang2017longitudinal}, and the emotional response to the patient's posts can affect the emotions of others \cite{qiu2011get,bui2016temporal,zhang2014does}, social media data can be used to (i) analyze and identify the characteristics of patients and (ii) help them have good eating habits, stable health condition with proper medication intake, and mental support \cite{de2015anorexia,he2016makes,wang2017detecting}.  
        \par Furthermore, investigating infectious diseases such as fever, influenza, and systemic inflammatory response syndrome (SIRS) were suggested to uncover key factors and subgroups, improve diagnosis accuracy, warn the public in advance, suggest appropriate prevention, and control strategies \cite{bodnar2014ground,tuarob2014ensemble, de2016hayfever,Alimova2017inps}. For instance, in \cite{chae2018predicting}, the authors addressed that infectious disease reports can be incomplete and delayed and used the search engines' data (both the health/medicine field search engine and the highest usage search engine), weather data from the Korea Meteorological Administration's weather information open portal, Twitter, and infectious disease data from the infectious disease web statistics system. It showed the possibility that deep learning can not only supplement current infectious disease surveillance systems but also predict trends in infectious disease, with immediate responses to minimize costs to society. Furthermore, \cite{covidment1} and \cite{covidment2} explained long-term pandemic situation might profoundly affect all aspects of society, not only the mortality rate, economic, but also the mental conditions of individuals. The authors in \cite{covidment1} and \cite{covidment2} used Twitter and online survey data to explore the effects of COVID-19 and how to address this circumstance.

\section{Challenges and Future Directions}
\par Artificial intelligence has given us an exploration of a new era. We reviewed how researchers implemented for different types of clinical data. Despite the notable advantages, there are some significant challenges.

    \subsection{Data}
    \par Medical data describe patients' health conditions over time. However, it is challenging to identify the true signals from the long-term context due to the complex associations among the clinical events. Data are high-dimensional, heterogeneous, temporal dependent, sparse, and irregular. Although the amount of datasets increases, still lack of labeled datasets and potential bias remain problems. Accordingly, data pre-processing and data credibility and integrity can also be thought of. 

        \subsubsection{Lack of Data and Labeled Data}
        \par Although there are no hard guidelines about the minimum number of training sets, more data can make stable and accurate models. However, in general, there is still no complete knowledge of the causes and progress of the disease, and one of the reasons is the lack of data. In particular, the number of patients is limited in a practical clinical scenario for rare diseases, certain age-related diseases, etc. 
\par Moreover, health informatics requires domain experts more than any other domain to label complex data and evaluate whether the model performs well and is practically applicable. Although labels generally help to have good performance of clinical outcomes or actual disease phenotypes, label acquisition is subjective and expensive.  
        \par The basis for achieving this goal is the availability of large amounts of data with well-structured data store system guidelines. Also, we need to attempt to label EHR data implicitly with unsupervised, semi-supervised, and transfer learning, as previous articles. In general, the first admission patient, disabled or transferred patients may be in worse health status and emergent circumstance, but with no information about medication allergy or any historical data. If we can use simple tests and calculate patient similarity to see each risk factor's potential, modifiable complications and crises will be reduced. 
	\par Furthermore, to train the target disease using different disease data, especially when the disease is class imbalanced, transfer learning, multi-task learning, reinforcement learning, and generalized algorithms can be considered. In addition, data generation and reconstruction can be other solutions besides incorporating expert knowledge from medical bibles, online medical encyclopedias, and medical journals. 
	\par Another potential data problems are data discrepancy and bias that current studies are based on retrospective data from hospitals that are readily available to share datasets. Because of patients' transferal, poor management of charts, and protocol tendency, we face missing value problems. Instead of simply cutting out data having missing value, researchers can try missing data imputation under no producing bias, matching cohorts, and reinforcement learning. Also, although we aim for generalized modeling through multi-center data, these studies still do not cover hospital data in developing countries and public hospitals overall. For these problems, imputation methods and matching cohorts can be used in the data pre-processing stage, and the attention mechanism can be applied to understand the results of models and derive an agreeable result with clinicians. Also, through reinforcement learning, not based only on the facts known, researchers can build models to provide medical intervention suggestions beyond data.

        \subsubsection{Data Pre-processing}
        \par Another important aspect to take into account when deep learning tools are employed is pre-processing. Encoding laboratory measurements by binary or low/medium/high or minimum/average/maximum ways, missing value interpolation, normalization or standardization were normally considered for pre-processing. Although it is a way to represent the data, especially when the data are high-dimensional, sparse, irregular, biased and multi-scale, none of DNN, CNN, and RNN models with one-hot encoding or AE or matrix/tensor factorization fully settled the problem. Thus, pre-processing, normalization or change of input domain, class balancing and hyperparameters of models are still a blind exploration process.
        \par In particular, considering temporal data, RNN/LSTM/GRU based models with vector-based inputs, as well as attention models, have already been used in previous studies and are expected to play a significant role toward better clinical deep architectures. However, we should point out that some patients with acute and chronic diseases have different time scales to investigate, and it can take a very long (5 years) time to track down for chronic diseases. Also, depending on the requirements, variables are measured with different timestamps (hourly, monthly, yearly time scale), and we need to understand how to handle those irregular time scale data. 
	\par Moreover, in previous studies, researchers completely cut out the patients with missing values, which induce bias in data. Also, researchers often did not report the whole steps of data pre-processing and did not evaluate whether the process could be justified without causing additional problems. However, these steps are necessary to evaluate the true performance of models.

        \subsubsection{Data Informativeness (high dimensionality, heterogeneity, multi-modality)}
        \par To cope with the lack of information and sparse, heterogeneous data and low dose radiation images, unsupervised learning for high-dimensionality and sparsity and multi-task learning for multi-modality have been proposed. Especially, in the case of multi-modality, these were studies that combined various clinical data types, such as medications and prescriptions in lab events from EHR, CT, and MRI from medical imaging. While deep learning research based on mixed data types is still ongoing, to the best of our knowledge, not so many previous literature provided attempts with different types of medical data, and the multi-modality related research is needed in the future with more reasons.
        \par First of all, even if we use long term medical records, sometimes it is not enough to represent the patients' status. It can be because of the time stamp to record, hospital discharge, or characteristics of data (ex. binary, low dose radiation image, short-term information provision data). In addition, even for the same CT or EHR, because hospitals use a variety of technologies, the collected data can be different based on CT equipment and basic or certified EHR systems. Furthermore, the same disease can appear very differently depending on clinicians in one institution when medical images are taken, EHRs are recorded, and clinical notes (abbreviations, ordering, writing style) are written. 
        \par With regard to outpatient monitoring and sharing of information on emergency and transferred patients, tracking the health status and summary for next hospital admission, it is necessary to obtain more information about patients to have a holistic representation of patient data. However, indeed, there are not much matched and structured data storing systems yet, as well as models. In addition, we need to investigate whether multi-task learning for different types of data is better than one task learning and if it is better, how deeply dividing types and how to combine the outcomes can be other questions. A primary attempt could be a divide-and-conquer or hierarchical approach or reinforcement learning to dealing with this mixed-type data to reduce dimensionality and multi-modality problems.
        
        \subsubsection{Data Credibility and Integrity}
        \par More than any other area, healthcare data are one of those heterogeneous, ambiguous, noisy, and incomplete but requires a lot of clean, well-structured data. For example, biosensor data and online data are in the spotlight. They can be a useful data source to continuously track health conditions even outside of clinical settings, extract people's feedback, sentiments, and detect abnormal vital signs. Furthermore, investigating infectious diseases can also improve diagnosis accuracy, warn the public in advance, and suggest appropriate prevention and management strategies. Accordingly, data credibility and integrity from biosensors, mobile applications, and online sources must be controlled.  
        \par First of all, if patients collect data and record their symptoms on websites and social media, it may not be worthwhile to use them in forecasting without proper instructions and control policies. At the point of data generation and collection, patients may not be able to collect consistent data, which may be affected by the environment. Patients with chronic illnesses may need to wear sensors and record their symptoms almost for the rest of their life, and it is difficult to expect consistently accurate data collection. Not only considering how it can be easy to wear devices, collect clear data, and combine clear and unclear data but also studying how we can educate patients most efficiently would be helpful. Besides, online community data can be written in unstructured languages such as misspellings, jokes, metaphors, slang, sarcasm, etc. Despite these challenges, there is a need for research that bridges the gap between all kinds of clinical information collected from hospitals and patients. And analyzing the data are expected to empower patients and clinicians to provide better health and life for clinicians and individuals.
        \par Second, the fact that patient signals are always detectable can be a privacy concern. People may not want to share data for a long time, which is why most of the research in this paper uses either a few de-identified publicly available hospital data or their own institution's privately available dataset. However, collecting and sharing all the possible information without condition, studying how to cover all the possible un-discovered data, and generating generalized models are the way to solve the fundamental data bias and model scalability problems. In addition, for mental illness patients, there is a limitation that patients who may and may not want to disclose their data have different starting points. In particular, when using the online community and social media, researchers should take into account the side effects. It can be much easier to try to use information and platform abusively in political and commercial purposes than any other data.

    \subsection{Model}
        \subsubsection{Model Interpretability and Reliability}
        \par Regardless of the data type, model credibility, interpretability, and how we can apply in practice will be another big challenge. The model or framework has to be accurate without overfitting and precisely interpretable to convince clinicians and patients to understand and apply the outcomes in practice. Especially when training data are small, noisy, and rare, a model can be easily fooled \cite{vogt2019precision}. Sometimes it seems that a patient has to have surgery with 90\% of certain diseases, but the patient is an unusual case, there may be no disease. However, opening the body can lead to high mortality due to complications, surgical burden, and the immune system. Because of concerns and assurances, there were studies using multi-modal learning and test normal images trained the model with images taken by PD patients, so that the model could be precise and generalized. The accuracy is important to convince users because it is related to cost, life-death problems, reliability, and others. At the same time, even if the prediction accuracy is superior to other algorithms, the interpretability is still important and should be taken care of to verify the results.
        \par Despite recent works on visualization with convolutional layers, clusters using t-SNE, word-cloud, similarity heatmaps, or attention mechanisms, deep learning models are often called as black boxes which are not interpretable. More than any other deterministic domains, in healthcare, such model interpretability is highly related to whether a model can be used practically for medications, hospital admissions, and operations, convincing both clinicians and patients. It will be a real hurdle if the model provider does not fully explain to the non-specialist why and how certain patients will have a certain disease with a certain probability on a certain date. Therefore, model credibility, interpretability, and application in practice should be equally important to healthcare issues. 
        
        \subsubsection{Model Feasibility and Security}
        \par Building and sharing models with other important research areas without leaking patient sensitive information will be an important issue in the future. If a patient agrees to share data with one clinical institution, but not publicly available to all institutions, our next question might be how to share data on what extent. In particular, deep learning-based systems for cloud computing-based biosensors and smartphone applications are growing, and we are emphasizing the importance of model interpretability. It can be a real concern if it is clearer to read the model with parameters, and some attacks violate the model and privacy. Therefore, we must consider research to protect the privacy of the models. 
        \par
\begin{figure}[!htbp]
    \centering
    \centerline{
    \includegraphics[scale=0.3]{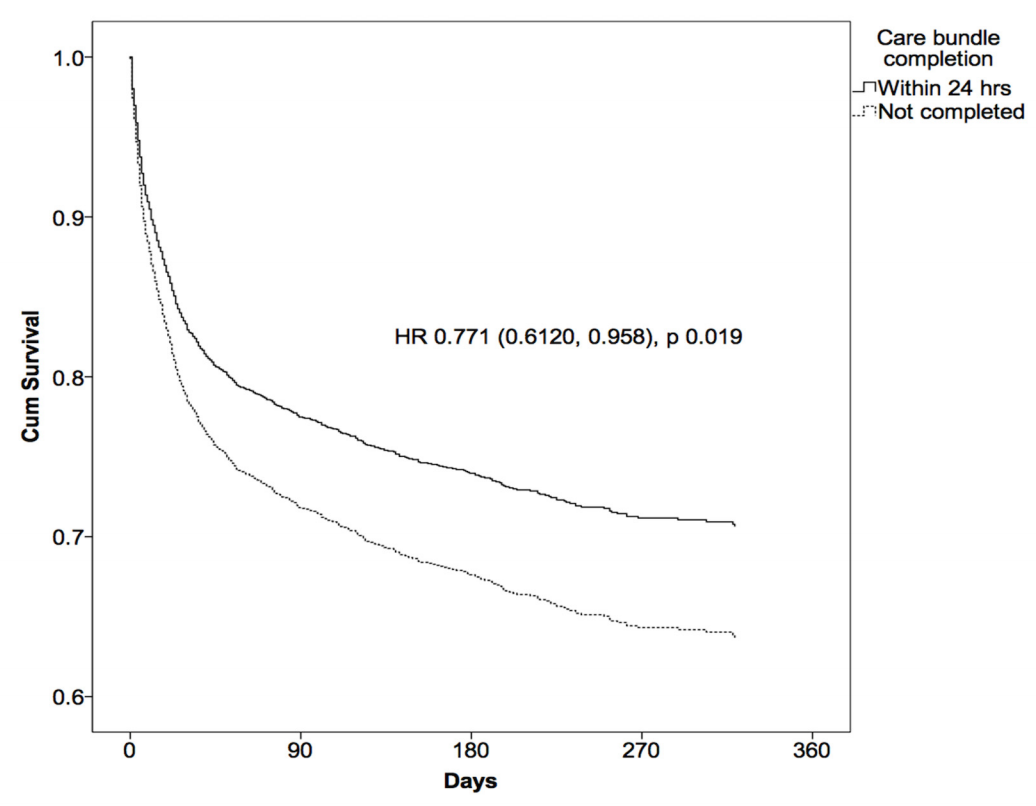}}
    \caption{Adjusted survival curve stratified by timing of completion of AKI care bundle \cite{kolhe2015impact}}
    \label{Img:Fig17}
    \end{figure}

 For cloud computing-based biosensors and smartphone applications, where and when model trained is another challenge. Training is a difficult and expensive process. Our biosensor and mobile app typically send requests to web services along with newly collected data, and the service stores data to train and replies with the prediction outcomes. However, some diseases progress quickly, and patients need immediate clinical care to avoid intensive care unit (ICU) admission (Fig.~\ref{Img:Fig17}). There have been studies on mobile devices, federated learning, reinforcement learning, and edge computing that focuses on bringing computing to the source of data closely. Researchers should study how to implement this system in healthcare as well as the development of algorithms for both acute and chronic cases. 

        \subsubsection{Model Scalability}
        \par Finally, we want to emphasize the opportunity to address the scalability of models. In most previous studies, either a few de-identified publicly available hospital datasets or their own institution's privately available datasets were used. However, patient health conditions and data at public hospitals or general practitioners (GP) clinics or disabled hospitals or developing countries' hospitals, which is not discovered yet, can be very different due to accessibility to the hospital and other reasons. In general, these hospitals may have less data information stored in the hospitals, but patients may be in an emergency with greater potential. We need to consider how our model with private hospitals or one hospital or one country can be extended for global use. In addition, clinicians provide medical care services based on protocols, and the protocols shift \cite{johnson2018comparative}, and the population changes over time \cite{google}. This is one of the reasons we consider an accurate but also generalized model to embrace time shifts.

\section{Conclusion}
\label{sec:conclusion}
\par To conclude, while there are several limitations, we believe that healthcare informatics with artificial intelligence can ultimately change human life. As more data become available, system supports, more researches are underway. Artificial intelligence can open up a new era of diagnosing disease, locating cancer, predicting the spread of infectious diseases, exploring new phenotypes, predicting strokes in outpatients, etc. This review provides insights into the future of personalized precision medicine and how to implement methods for clinical data to support better health and life.

\label{sec:references}
\bibliography{template}
\bibliographystyle{unsrt}



\end{document}